\documentclass[10pt,twocolumn,letterpaper]{article}
\PassOptionsToPackage{table}{xcolor}
\usepackage{iccv}              

\definecolor{iccvblue}{rgb}{0.21,0.49,0.74}
\usepackage[pagebackref,breaklinks,colorlinks,allcolors=iccvblue]{hyperref}

\usepackage{array, multirow, bm}
\usepackage{amssymb}
\usepackage{booktabs}       

\usepackage{algorithm}
\usepackage{algorithmic}

\usepackage{graphicx}

\graphicspath{{Figure/}}
\newcommand{\tb}[1]{\textbf{#1}}
\newcommand{\ti}[1]{\textcolor{Blue}{#1}}
\newcommand{\ul}{\underline}
\newcommand{\std}[1]{\scriptsize{$\pm$#1}}

\def\paperID{8690} 
\def\confName{ICCV}
\def\confYear{2025}

\title{GenHPE: Generative Counterfactuals for 3D Human Pose Estimation with Radio Frequency Signals}

\author{Shuokang Huang\quad Julie A. McCann\\
Imperial College London\\
{\tt\small s.huang21@imperial.ac.uk\quad j.mccann@imperial.ac.uk}
}

\begin{document}
	\maketitle
	\begin{abstract}
		%
		Human pose estimation (HPE) detects the positions of human body joints for various applications.
		%
		Compared to using cameras, HPE using radio frequency (RF) signals is non-intrusive and more robust to adverse conditions, exploiting the signal variations caused by human interference.
		%
		However, existing studies focus on single-domain HPE confined by domain-specific confounders, which cannot generalize to new domains and result in diminished HPE performance.
		Specifically, the signal variations caused by different human body parts are entangled, containing subject-specific confounders.
		RF signals are also intertwined with environmental noise, involving environment-specific confounders.
		%
		%
		%
		%
		In this paper, we propose GenHPE, a 3D HPE approach that generates counterfactual RF signals to eliminate domain-specific confounders.
		GenHPE trains generative models conditioned on human skeleton labels, learning how human body parts and confounders interfere with RF signals.
		We manipulate skeleton labels (\textit{i.e.}, removing body parts) as counterfactual conditions for generative models to synthesize counterfactual RF signals. 
		%
		The differences between counterfactual signals approximately eliminate domain-specific confounders and regularize an encoder-decoder model to learn domain-independent representations.
		Such representations help GenHPE generalize to new subjects/environments for cross-domain 3D HPE.
		%
		%
		%
		%
		%
		%
		We evaluate GenHPE on three public datasets from WiFi, ultra-wideband, and millimeter wave.
		Experimental results show that GenHPE outperforms state-of-the-art methods and reduces estimation errors by up to 52.2mm for cross-subject HPE and 10.6mm for cross-environment HPE.
		%
	\end{abstract}

	\section{Introduction}
		\label{sec_introduction}
		\begin{figure}[t]
			\centering
			\begin{subfigure}{\linewidth}
				\centering
				\includegraphics[width=1\linewidth]{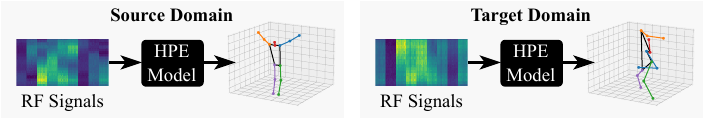}
				\caption{Existing 3D HPE methods with RF signals.}
				\label{figure_genhpe_overview_0}
				\vspace{1.5mm}
			\end{subfigure}
			\begin{subfigure}{\linewidth}
				\centering
				\includegraphics[width=1\linewidth]{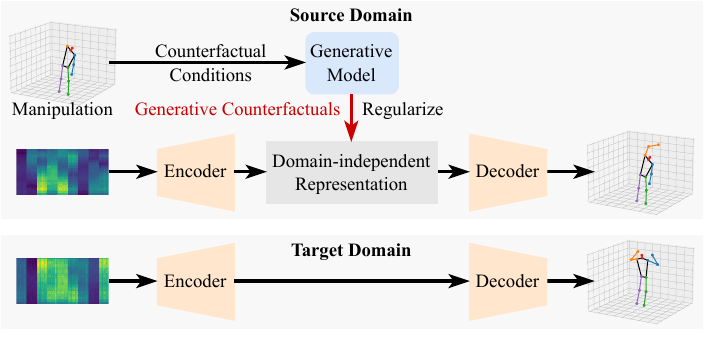}
				\caption{Generative counterfactuals for 3D HPE with RF signals.}
				\label{figure_genhpe_overview_1}
			\end{subfigure}
			\caption{%
				Comparison between (a) existing 3D HPE methods and (b) GenHPE in cross-domain scenarios.
				Existing methods train HPE models confined by domain-specific confounders and suffer from severe performance degradation in new domains.
				%
				GenHPE introduces a generative model to help eliminate domain-specific confounders and regularizes an encoder-decoder model to learn domain-independent representations for cross-domain 3D HPE.
			}
			\vspace{-1mm}
			\label{figure_genhpe_overview}
		\end{figure}
		
		%
		Human pose estimation (HPE) is a fundamental technique to predict the coordinates of human body joints for diverse applications, such as healthcare \cite{app_healthcare}, autonomous driving \cite{app_auto_drive}, and augmented/virtual reality \cite{app_avr}.
		%
		Recent works have explored various signal sources for HPE, including cameras \cite{related_hrnet,related_poseirm,related_selfpose3d}, wearable sensors \cite{related_sensor_0,related_sensor_1,related_sensor_2}, and radio frequency (RF) \cite{method_piw_3d,data_mmfi,data_operanet}.
		%
		In contrast to cameras and wearable sensors, RF-based HPE does not visually record human subjects or attach sensors to them, which is less intrusive and offers a degree of privacy protection \cite{survey_non_intrusive}, making it widely applicable.
		%
		Moreover, most cameras are highly sensitive to occlusion or darkness, whereas RF signals can continue detection even in these adverse conditions \cite{related_wall_0,related_wall_1,related_env_robust}.
		
		%
		The principle of using RF signals for HPE is that human bodies interfere with RF signals and lead to signal variations \cite{related_rf_if, related_wifi_if}.
		%
		Such variations implicitly contain human features and therefore can be exploited for HPE \cite{related_rf_pose}.
		%
		For example, WiFi devices record signal variations in Channel State Information (CSI), which can be utilized for WiFi-based HPE \cite{data_mmfi} with Recurrent Neural Networks (RNNs) \cite{method_wipose}, Convolutional Neural Networks (CNNs) \cite{method_wpnet}, Transformers \cite{method_wpformer,method_piw_3d}, \textit{etc}.
		%
		Similarly, ultra-wideband (UWB) radars have been used for HPE with CNNs \cite{related_uwb_0,related_uwb_1,data_operanet}.
		%
		Existing methods have also explored millimeter wave (mmWave) signals for HPE \cite{data_mmfi} based on Transformers \cite{method_pt,method_poseformer}, Graph Convolution Networks (GCNs)\cite{method_mmdiff}, and diffusion models \cite{method_diffpose,method_mmdiff}.
		%
		
		%
		Despite remarkable advancements, current RF-based HPE methods are limited to single domains and struggle to generalize in cross-domain scenarios \cite{survey_cross_domain}.
		%
		Specifically, single-domain HPE methods are inevitably biased towards domain-specific confounders, which do not exist in new domains and hamper the performance of cross-domain HPE.
		For example, compared to single-domain scenarios, the HPE errors of PiW-3D \cite{method_piw_3d} increase by 39.9mm in cross-subject scenarios and 534.7mm in cross-environment scenarios. 
		%
		Since all body parts of human subjects interfere with RF signals simultaneously, their corresponding signal variations entangle and include subject-specific confounders.
		%
		Environmental noise also induces environment-specific confounders in RF signals.
		%
		These confounders hinder models from generalizing to new subjects/environments and motivate us to investigate:

		\textit{\textbf{How to eliminate domain-specific confounders in RF signals and represent the signal variations caused solely by each human body part for cross-domain HPE?}}
		
		%
		The core challenge is that, in reality, it is impossible to separate body parts from a living person and study each body part individually.
		Meanwhile, excessive environmental noise makes it difficult to eliminate environment-specific confounders.
		To tackle these challenges, we propose GenHPE, a 3D HPE method using generative models to learn how body parts and confounders interfere with RF signals.
		%
		GenHPE trains generative models to synthesize RF signals conditioned on ground-truth human skeleton labels.
		After the training, we manipulate skeleton labels (\textit{i.e.}, removing human body parts iteratively) as counterfactual\footnotemark conditions, under which generative models synthesize counterfactual RF signals.
		A series of counterfactual signals simulate the variations if each body part is ``removed''.
		%
		We calculate the differences between these signals to eliminate domain-specific confounders and aggregate these differences to approximate signals related to all body parts yet independent of domains.
		%
		%
		%
		%
		%
		The aggregated signals regularize an encoder to learn domain-independent representations, which are fed to a decoder for HPE.
		%
		Since the encoder extracts representations independent of domains, the encoder-decoder model can generalize to new subjects/environments (target domains) for cross-domain 3D HPE with RF signals (Figure \ref{figure_genhpe_overview}).
		\footnotetext{We use the term ``counterfactual'' to describe what cannot happen in reality (\textit{e.g.}, a living person without head).}
		We summarize our contributions as follows:
		\begin{enumerate}
			\item[\textbullet]{%
				We propose GenHPE, to the best of our knowledge, the first work of domain generalization for RF-based 3D HPE and the first attempt using generative models to help RF-based models generalize across domains.
				%
			}
			\item[\textbullet]{%
				%
				We devise a difference-based scheme to remove domain-specific confounders with counterfactual RF signals and regularize an encoder-decoder model to learn domain-independent representations for cross-domain HPE.
			}
			\item[\textbullet]{%
				Experimentally, we apply three generative models (\textit{i.e.}, two diffusion models and a generative adversarial network) in GenHPE for three RF sources (\textit{i.e.}, WiFi, UWB, and mmWave) on public datasets.
				%
				Compared with state-of-the-art methods, GenHPE remarkably reduces 3D HPE errors in cross-subject/environment scenarios.
			}
		\end{enumerate}

	\begin{figure*}[t]
		\centering
		\includegraphics[width=1.0\linewidth]{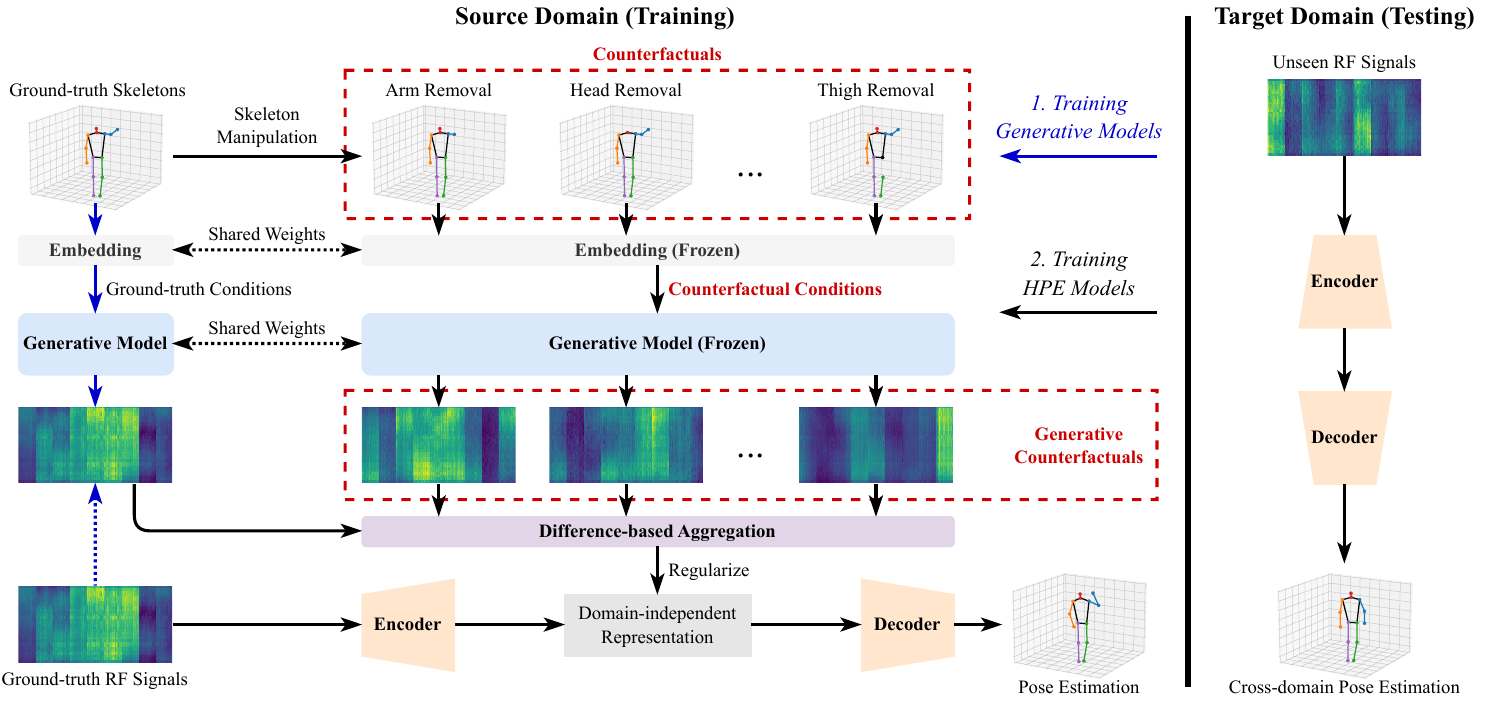}
		\caption{Overview of the proposed GenHPE.
			\textcolor[HTML]{0000CC}{\textbf{\textit{Step 1}}}: In the source domain, we train a generative model using ground-truth RF signals and skeleton labels. 
			\textbf{\textit{Step 2}}: We manipulate skeleton labels by removing each body part (\textit{e.g.}, arms, head, and thighs) and use them as counterfactual conditions for the generative model to synthesize counterfactual RF signals.
			%
			The differences between counterfactual signals eliminate domain-specific confounders and are aggregated to approximate signals only related to body parts, regularizing an encoder-decoder model to learn domain-independent representations. 
			Such representations generalize to both source and target domains for HPE.
			%
		}
		\label{figure_genhpe}
	\end{figure*}

	\section{Related Work}
		\label{sec_related}
		\subsection{3D Human Pose Estimation with RF Signals}
			Recent years have witnessed the rapid growth of HPE using RF signals, thanks to its lower intrusiveness and its robustness to challenging situations. 
			Existing approaches have utilized RF signals from WiFi, UWB, and mmWave.

			\textit{WiFi-based 3D HPE.}
			WiFi has been studied for 3D HPE because WiFi signals are ubiquitous for device-free sensing without dedicated deployments.
			WiPose \cite{method_wipose} and GoPose \cite{related_wifi_gopose} combine CNNs and RNNs to estimate human poses from WiFi CSI, showing the practicability of WiFi-based 3D HPE.
			Winect \cite{related_wifi_winect} and WPNet \cite{method_wpnet} further establish ResNet models \cite{related_resnet} to improve 3D HPE performance using WiFi.
			Given the success of Transformers \cite{related_attention}, WPFormer \cite{method_wpformer} and PiW-3D \cite{method_piw_3d} leverage the attention mechanism to achieve state-of-the-art WiFi-based 3D HPE.
			Recent works have further explored the use of WiFi for hand skeleton estimation \cite{related_wifi_hand}, human mesh construction \cite{related_wifi_mesh}, \textit{etc}.

			\textit{UWB-based 3D HPE.}
			UWB-based human sensing, which is less popular due to the need for additional devices, has used higher-frequency signals for hand gesture recognition \cite{related_uwb_0} and respiration monitoring \cite{related_uwb_1}.
			OPERAnet \cite{data_operanet} has demonstrated its feasibility for 3D HPE with UWB signals.

			\textit{mmWave-based 3D HPE.}
			mmWave radars gain increasing popularity for higher-resolution 3D HPE using higher signal frequencies (over 30 GHz) than WiFi (2.4/5 GHz) and UWB (3.1$\sim$10.6 GHz) \cite{related_mmwave,data_mmfi}.
			Given the 3D point clouds from mmWave radars, Point Transformer \cite{method_pt} and PoseFormer \cite{method_poseformer} develop Transformer-based methods for 3D HPE.
			DiffPose \cite{method_diffpose} and mmDiff\cite{method_mmdiff} adopt diffusion models for more accurate estimation with mmWave signals.

			Despite their promising performance, the above 3D HPE methods are limited to single domains and fall short in domain generalization.
			%
			Recent works have examined cross-domain human sensing with RF signals, such as RF-Net \cite{related_rfnet} which studies environment adaptation for human activity recognition, as well as Widar3.0 \cite{related_widar3} and AirFi \cite{related_airfi} for environment-independent gesture recognition.
			%
			%
			%
			%
			However, they only focus on solving coarse-grained classification problems, with finite and discrete outputs.
			Cross-domain 3D HPE with RF signals is finer-grained, requiring infinite and continuous outputs, and remains unexplored.
			Furthermore, domain adaptation methods \cite{related_rfnet} need data from new (target) domains for adaptations, requiring more effort for data collection.
			%
			%
			Our method aims to directly learn domain-independent representations which can generalize to new subjects/environments without domain adaptation.

		\subsection{Generative Counterfactuals}
			%
			Generative models have exhibited remarkable ability for data synthesis.
			%
			Based on adversarial learning, Generative Adversarial Networks (GANs) \cite{generative_gan} can perform high-quality data synthesis, deriving Conditional GANs (CGANs) \cite{generative_cgan}, Wasserstein GAN (WGANs) \cite{generative_wgan}, \textit{etc}.
			Compared to GANs using one-step synthesis from noise to data, diffusion models have been introduced to synthesize data by multiple-step denoising to outperform GANs, including Denoising Diffusion Probabilistic Models (DDPMs) \cite{generative_ddpm} and Denoising Diffusion Implicit Models (DDIMs) \cite{generative_ddim}.
			Specifically, generative models contribute to counterfactual synthesis for various tasks, such as model explanation \cite{related_explain_0,related_explain_1} and data augmentation \cite{related_augment_0,related_augment_1,related_diffar}.

			Recently, generative models have served to synthesize counterfactual features as interventions for representation learning.
			For example, GANs have been employed to intervene an HPE model for advanced generalization ability \cite{related_intervene_0}.
			GaitGCI \cite{related_intervene_1} enhances camera-based gait recognition by maximizing the likelihood difference between ground-truth/synthetic features.
			In this paper, we pioneer the use of generative counterfactuals to enhance domain generalization for RF-based 3D HPE. 

	\section{GenHPE}
		\label{sec_method}
		%
		We outline GenHPE in Figure \ref{figure_genhpe}, which learns domain-independent representations with generative counterfactuals and difference-based aggregation for cross-domain HPE.
		\subsection{Preliminary}
			Given a raw RF signal sample $\bm{x}$, 3D HPE aims to predict the ground-truth coordinates of $N$ human body joints $\bm{y}\in\mathbb{R}^{N\times 3}$, where $\bm{y}=\{\bm{y}_1, \bm{y}_2, ..., \bm{y}_N\}$ and $\bm{y}_n\in\mathbb{R}^{3}$.
			A 3D HPE model $f(\cdot)$ takes $\bm{x}$ as input to output $\hat{\bm{y}}=f(\bm{x})$, and we aim to minimize the error between $\hat{\bm{y}}$ and $\bm{y}$.
			From the aspect of probability distributions, $f(\cdot)$ models a distribution $\hat{\bm{y}}\sim p_f(\hat{\bm{y}}|\bm{x})$ to estimate the ground-truth $\bm{y}\sim p_{\mathrm{gt}}(\bm{y}|\bm{x})$.

			\textit{Domain-specific Confounders.}
			We can use a joint distribution $p_{\mathrm{gt}}(\bm{x}^{\bm{h}},\bm{x}^d)$ to represent RF signal samples, where $\bm{x}^{\bm{h}}$ represents human-related signal variations and $\bm{x}^d$ represents domain-specific confounders.
			Therefore, the 3D HPE model actually learns a distribution $\hat{\bm{y}}\sim p_f\left(\hat{\bm{y}}|\bm{x}^{\bm{h}},\bm{x}^d\right)$ including the confounders $\bm{x}^d$.
			In the source domain $d_0$, existing works can model $p_f(\hat{\bm{y}}|\bm{x}^{\bm{h}},\bm{x}^{d_0})$ to estimate the ground-truth $p_{\mathrm{gt}}(\bm{y}|\bm{x}^{\bm{h}},\bm{x}^{d_0})$ with minimized errors.
			However, if we apply $f(\cdot)$ in a new (target) domain $d_1$, there exists a gap between $p_f(\hat{\bm{y}}|\bm{x}^{\bm{h}},\bm{x}^{d_0})$ and $p_{\mathrm{gt}}(\bm{y}|\bm{x}^{\bm{h}},\bm{x}^{d_1})$ due to the difference between confounders $\bm{x}^{d_0}$ and $\bm{x}^{d_1}$.
			To solve these issues in cross-domain 3D HPE, we study how to \textit{approximately eliminate domain-specific confounders $\bm{x}^d$ using the differences between generative counterfactual RF signals}.

		\subsection{Generative Counterfactual Synthesis}
			%
			We transform human body joints $\bm{y}$ into skeleton vectors $\bm{h}$, whose embeddings $\bm{c}$ are used as conditions to train a generative model $g(\cdot)$ for the synthesis of RF signals $\hat{\bm{x}}$.
			With the trained generative model $g(\cdot)$, we manipulate skeleton vectors as counterfactual conditions for $g(\cdot)$ to synthesize RF signals $\bar{\bm{x}}^{\bm{h}_k}$ which removes the impact of the $k$-th skeleton vector $\bm{h}_k$.
			%
			Using the difference between $\hat{\bm{x}}$ and $\bar{\bm{x}}^{\bm{h}_k}$, we draw out $\bm{r}_k$ to eliminate domain-specific confounders and represent the signals \textit{only} related to $\bm{h}_k$.
			Given $K$ skeleton vectors, we aggregate all $\bm{r}_k$ for $k\in[1,...,K]$ as $\bm{r}$ to approximate skeleton-related yet domain-independent signals.

			\noindent\textbf{Skeleton Embedding.}
				Typically, HPE labels are in the form of discrete human body joint descriptions, but what really interfere with RF signals are different human body parts.
				Therefore, we embody body parts with skeleton embeddings and use them for generative models to synthesize RF signals.
				We adopt a skeleton map $\bm{s}\in\mathbb{R}^{K\times 2}$ to indicate $K$ bones in each skeleton, where the $k$-th bone is connected between the joints indexed by $\bm{s}_{k, 1}$ and $\bm{s}_{k, 2}$.
				Given $N$ human body joints $\bm{y}\in\mathbb{R}^{N\times 3}$, the corresponding $k$-th bone can be represented by $\bm{h}_k= \left(\bm{y}_{\bm{s}_{k, 1}}, \bm{y}_{\bm{s}_{k, 2}}\right)\in\mathbb{R}^{2\times 3}$.
				Accordingly, we can transform $\bm{y}$ into $K$ skeleton vectors as:
				\begin{equation}
					\label{equ_skeleton_vector}
					\bm{h} = \left\{ \bm{h}_k\ |\ k\in \left[1,2,...,K\right] \right\}\in\mathbb{R}^{K\times 2\times 3}.
				\end{equation}
				We project skeleton vectors $\bm{h}$ into an embedding space using a multilayer perceptron $\varphi(\cdot)$ to derive a skeleton embedding vector $\bm{c} = \varphi\left(\bm{h}\right)$.
				%
				%
				We equip $\varphi(\cdot)$ with two linear layers, and each linear layer is followed by a Sigmoid Linear Unit (SiLU) activation function.

			\noindent\textbf{Conditional Generative Models.}
				Diverse conditional generative models have been proposed to synthesize specific samples, including conditional diffusion models \cite{generative_cdm} and CGANs \cite{generative_cgan}.
				We employ generative models conditioned on skeleton embeddings $\bm{c}$ to synthesize RF signals $\hat{\bm{x}}$.
				Specifically, most generative models perform synthesis from noise $\bm{z}$ to data.
				Therefore, we formulate a conditional generative model $g\left(\cdot\right)$ for RF signal synthesis as: 
				\begin{equation}
					\label{equ_con_gen_0}
					\hat{\bm{x}} = g\left(\bm{z}, \bm{c}\right) = g\left(\bm{z}, \varphi\left(\bm{h}\right)\right).
				\end{equation}
				%
				%
				From the aspect of probability distributions, $g\left(\cdot\right)$ essentially learns a generative distribution $p_g$ as:
				\begin{equation}
					\label{equ_con_gen_1}
					\begin{aligned}
						\hspace{-1mm}\hat{\bm{x}}\sim p_g\left(\hat{\bm{x}}\ |\ \bm{h}\right) &= p_g\left(\hat{\bm{x}}^{\bm{h}_1}, ..., \hat{\bm{x}}^{\bm{h}_K}, \hat{\bm{x}}^{d} \ |\ \bm{h}_1, ..., \bm{h}_K\right) \\
						&= p_g\left(\hat{\bm{x}}^{\bm{h}_{1:K}}, \hat{\bm{x}}^{d} \ |\ \bm{h}_{1:K}\right),
					\end{aligned}
				\end{equation}
				where $\hat{\bm{x}}^{\bm{h}_k}$ represents the signal variations caused by $\bm{h}_k$, and $\hat{\bm{x}}^{d}$ denotes the learned domain-specific confounders.
				Such a generative distribution can simulate how each body part $\bm{h}_k$ interferes with RF signals, and we need to manipulate skeletons $\bm{h}$ to further eliminate the confounders $\hat{\bm{x}}^{d}$.

			\noindent\textbf{Counterfactual Skeleton Manipulation.}
				In reality, it is impossible to separate different body parts from a living person for study.
				However, we can manipulate skeleton vectors $\bm{h}$ by removing each vector $\bm{h}_k$.
				A manipulation function $m_k(\cdot)$ removes the $k$-th vector from $\bm{h}$ to yield counterfactual skeleton vectors $\bar{\bm{h}}_k$ as:
				\begin{equation}
					\label{equ_ske_mani_0}
					\bar{\bm{h}}_k=m_k\left(\bm{h}\right)=\left\{\bm{h}_i\ |\ i\in[1,2,...,K]\ \mathrm{and}\ i\neq k\right\}.
				\end{equation}
				We use such counterfactuals as conditions for generative models to synthesize RF signals.
				Specifically, we project $\bar{\bm{h}}_k$ into the embedding space as $\bar{\bm{c}}_k=\varphi\left(\bar{\bm{h}}_k\right)$ and feed $\bar{\bm{c}}_k$ into $g(\cdot)$ to sample counterfactual RF signals $\bar{\bm{x}}^{\bm{h}_k}$ as:
				\begin{equation}
					\label{equ_ske_mani_1}
					\bar{\bm{x}}^{\bm{h}_k} = g\left(\bm{z}, \bar{\bm{c}}_k\right) = g\left(\bm{z}, \varphi\left(\bar{\bm{h}}_k\right)\right),
				\end{equation}
				where the parameters of $\varphi(\cdot)$ and $g(\cdot)$ are frozen in this phase.
				Such a counterfactual synthesis approximately samples RF signals from the following distribution:
				\begin{equation}
					\label{equ_ske_mani_2}
					\bar{\bm{x}}^{\bm{h}_k}\sim p_g\left(\hat{\bm{x}}^{\bm{h}_{1:k-1}}, \hat{\bm{x}}^{\bm{h}_{k+1,K}}, \hat{\bm{x}}^{d} \ |\ \bm{h}_{1:k-1}, \bm{h}_{k+1:K}\right),
				\end{equation}
				which represents the signal variations if the $k$-th human body part is ``removed''. 
				In practice, $\varphi(\cdot)$ and $g(\cdot)$ require inputs with unchanged dimensions during training and sampling, and thus we can set the values of $\bm{h}_k$ as $0$ as removal. 
				Note that the domain-specific confounders $\hat{\bm{x}}^{d}$ still exist in $\bar{\bm{x}}^{\bm{h}_k}$, and we devise a difference-based aggregation scheme to approximately eliminate these confounders.

			\noindent\textbf{Difference-based Aggregation.}
				%
				Comparing Equation (\ref{equ_con_gen_1}) and (\ref{equ_ske_mani_2}),
				$\hat{\bm{x}}$ represents the signal variations caused by the confounders $\hat{\bm{x}}^{d}$ and the complete skeleton $\bm{h}_{1:K}$, while $\bar{\bm{x}}^{\bm{h}_k}$ represents the signal variations without the impact of $\bm{h}_k$.
				Intuitively, we can \textit{represent the signal variations \textbf{only} caused by $\bm{h}_k$} based on the difference between $\hat{\bm{x}}$ and $\bar{\bm{x}}^{\bm{h}_k}$.
				We apply a distance function $\phi(\cdot)$ to calculate such a difference $\bm{r}_k$, 
				%
				%
				which approximates $\hat{\bm{x}}^{\bm{h}_k}$ only related to $\bm{h}_k$ as:
				\begin{equation}
					\label{equ_diff_1}
					\bm{r}_k = \phi\left(\hat{\bm{x}},\ \bar{\bm{x}}^{\bm{h}_k}\right) \approx \hat{\bm{x}}^{\bm{h}_k} \sim p_g\left(\hat{\bm{x}}^{\bm{h}_k}\ |\ \bm{h}_k\right).
				\end{equation}
				In practice, we can simply implement the distance function with element-wise subtraction $\phi(a, b) = a-b$.
				Such a difference removes environmental noise and disentangles concurrent signal variations caused by multiple body parts.

				Therefore, $\bm{r}_k$ is independent of domains (\textit{i.e.}, environments or subjects) and only related to a single body part $\bm{h}_k$.
				%
				%
				%
				We can calculate $\bm{r}_k$ for $k\in[1,2,...,K]$ iteratively and aggregate them as $\bm{r}$.
				Accordingly, \textit{$\bm{r}$ represents the impacts of all body parts and is also independent of domains.}
				Note that RF signals from various sources have different meanings.
				For example, WiFi and UWB collect CSI data which are multichannel sequences, while mmWave collects radar point clouds in 3D space.
				We concatenate $\bm{r}_k$ for $k\in[1,2,...,K]$ in a new dimension and apply a linear layer to aggregate them as $\bm{r} = \mathrm{linear}\left( \mathrm{concat} \left(\bm{r}_1, \bm{r}_2, ..., \bm{r}_K\right) \right)$.
				\ul{Appendix \ref{appendix_exp_dataset}} provides more details about data dimensions.
				%
		
		%
		\subsection{Domain-independent Representation Learning}
			After eliminating domain-specific confounders, $\bm{r}$ is independent of domains and represents the signal variations caused solely by human body parts, which can approximate skeleton-related yet domain-independent signals as:
			%
			\begin{equation}
				\label{equ_drl_0}
				\bm{r} \approx \hat{\bm{x}}^{\bm{h}}\sim p_g\left(\hat{\bm{x}}^{\bm{h}_{1:K}}\ |\ \bm{h}_{1:K}\right).
			\end{equation}
			%
			Compared to Equation (\ref{equ_con_gen_1}), $\bm{r}$ eliminates $\bm{x}^d$ and thus can regularize models to generalize across domains.
			To this end, we build up an encoder-decoder 3D HPE model.
			%
			%
			The encoder $f_{\mathrm{en}}(\cdot)$ takes ground-truth RF signal samples $\bm{x}$ as input and learns domain-independent representations $\bm{v}=f_{\mathrm{en}}(\bm{x})$ with the same dimensions as $\bm{r}$.
			The decoder $f_{\mathrm{de}}(\cdot)$ further projects $\bm{v}$ into the 3D space for HPE as:
			\begin{equation}
				\label{equ_drl_1}
				\hat{\bm{y}} = f_{\mathrm{de}}\left(\bm{v}\right) = f_{\mathrm{de}}\left(f_{\mathrm{en}}\left(\bm{x}\right)\right).
			\end{equation}
			%
			%
			We optimize the encoder-decoder model using a pose estimation loss $\mathcal{L}_{\mathrm{pe}}$ between $\hat{\bm{y}}$ and $\bm{y}$.
			Meanwhile, we apply counterfactual regularization $\mathcal{L}_{\mathrm{cr}}$ based on the likelihood between $\bm{v}$ and $\bm{r}$, enforcing the encoder to learn domain-independent representations.
			The overall loss $\mathcal{L}$ of GenHPE can be formulated as:
			%
			\begin{equation}
				\label{equ_loss_sum}
				\mathcal{L} = \mathcal{L}_{\mathrm{pe}}\left(\hat{\bm{y}}, \bm{y}\right) + \lambda\cdot\mathcal{L}_{\mathrm{cr}}\left(\bm{v}, \bm{r}\right),
			\end{equation}
			where $\lambda$ is a ratio to balance pose estimation loss and counterfactual regularization.
			%
			Since the encoder has learned representations that are independent of domains, the trained encoder-decoder model can generalize to target domains  (new subjects/environments) without domain adaptation.

		\subsection{Overall Training in GenHPE}
			\noindent\textbf{Training Generative Models.}
			GenHPE explores three generative models that have been widely discussed and time-tested for high-quality data synthesis, including DDPMs \cite{generative_ddpm,generative_cdm}, DDIMs \cite{generative_ddim}, and CGANs \cite{generative_cgan,generative_wgan}.
			More details about generative models are presented in \ul{Appendix \ref{appendix_gen_model}}.

			\textit{Conditional DDPM.}
				%
				%
				%
				DDPM \cite{generative_ddpm} introduces a forward process where a fixed $T$-step Markov chain converts raw data $\bm{x}_0$ to Gaussian noise $\bm{x}_T$.
				Each step in the forward process is a fixed Gaussian transition $q\left(\bm{x}_{t}|\bm{x}_{t-1}\right)$. 
				%
				%
				Conversely, DDPM devises a reverse process using $T$ steps to synthesize $\hat{\bm{x}}_0$ from noise $\hat{\bm{x}}_T$.
				Each step in the reverse process is a learnable Gaussian transition $p_\theta\left(\hat{\bm{x}}_{t-1}|\hat{\bm{x}}_{t}\right)$. 
				Given the data distribution $q\left(\bm{x}_0|\bm{c}\right)$ under conditions $\bm{c}$, conditional DDPM \cite{generative_cdm} aims to maximize the likelihood $\mathbb{E}\left[p_\theta\left(\bm{x}_0|\bm{c}\right)\right]$.
				%
				In practice, conditional DDPM minimizes the negative log likelihood $\mathbb{E}\left[-\log p_\theta\left(\bm{x}_0|\bm{c}\right)\right]$ by minimizing the divergence between
				$p_\theta\left(\hat{\bm{x}}_{t-1}|\hat{\bm{x}}_{t}, \bm{c}\right)$ and
				$q\left(\bm{x}_{t-1}|\bm{x}_t,\bm{x}_0, \bm{c}\right)$.
				Since $q\left(\bm{x}_{t-1}|\bm{x}_t,\bm{x}_0, \bm{c}\right)$ is a Gaussian distribution with a fixed mean $\tilde{\bm{\mu}}_t$, $p_\theta\left(\hat{\bm{x}}_{t-1}|\hat{\bm{x}}_{t}, \bm{c}\right)$ can learn $\tilde{\bm{\mu}}_t$ with a parameterized mean $\bm{\mu}_\theta$.
				%
				%
				Such a process can be simplified as training a model $\bm{\epsilon}_\theta\left(\cdot\right)$ to estimate $\bm{\epsilon}\sim \mathcal{N}\left(\textbf{0}, \mathbf{I}\right)$ as:
				%
				%
				\begin{equation}
					\label{equ_ddpm_train}
					\hspace{-2mm}\mathcal{L}\left(\theta\right) := \mathbb{E}_{\bm{x}_0}\hspace{-1mm}\left[ \left\| \bm{\epsilon} - \bm{\epsilon}_\theta\left(\sqrt{\bar{\alpha}_t} \bm{x}_0 + \sqrt{1-\bar{\alpha}_t} \bm{\epsilon}, t, \bm{c} \right)\right\|^2 \right]\hspace{-0.5mm},\hspace{-1.5mm}
				\end{equation}
				After training $\bm{\epsilon}_\theta\left(\cdot\right)$, conditional DDPM synthesizes $\hat{\bm{x}}_0$ from $\hat{\bm{x}}_T\sim \mathcal{N}\left(\textbf{0}, \mathbf{I}\right)$ by iterating $t\in[T,...,1]$ as:
				\begin{equation}
					\label{equ_ddpm_synthesis}
					\hat{\bm{x}}_{t-1} = \frac{1}{\sqrt{\alpha_t}} \left( \hat{\bm{x}}_t - \frac{\beta_t}{\sqrt{1-\bar{\alpha}_t}} \bm{\epsilon}_\theta \left(\hat{\bm{x}}_t, t, \bm{c}\right) \right) + \sigma_t\bm{z}.
				\end{equation}
				%
				%
				DDPM needs many steps (\textit{e.g.}, $T$ = 1000) for high-quality synthesis, which is time-consuming.
				Hence, DDIM \cite{generative_ddim} focuses on faster synthesis with fewer steps.

			\textit{Conditional DDIM.}
				The training of DDIM \cite{generative_ddim} is the same as that of DDPM, and thus we can reused the trained $\bm{\epsilon}_\theta\left(\cdot\right)$ in DDPM.
				DDIM uses $T_S$ synthesis steps which are flexible and can be fewer than $T$ in training $\bm{\epsilon}_\theta\left(\cdot\right)$.
				Given a trained $\bm{\epsilon}_\theta\left(\cdot\right)$, conditional DDIM synthesizes $\hat{\bm{x}}_0$ from $\hat{\bm{x}}_T\sim \mathcal{N}\left(\textbf{0}, \mathbf{I}\right)$  with $t\in[T_S,...,1]$ as\footnotemark:
				\begin{equation}
					\label{equ_ddim_synthesis}
					\begin{aligned}
						\hat{\bm{x}}_{t-1} = &\sqrt{\bar{\alpha}_{t-1}} \left(\frac{\hat{\bm{x}}_{t}-\sqrt{1-\bar{\alpha}_t}\cdot\bm{\epsilon}_\theta\left(\hat{\bm{x}}_t,t,\bm{c}\right)}{\sqrt{\bar{\alpha}_t}}\right)\\
						&+ \sqrt{1-\bar{\alpha}_{t-1}-\sigma_t^2}\cdot\bm{\epsilon}_\theta\left(\hat{\bm{x}}_t,t,\bm{c}\right) + \sigma_t \bm{z},
					\end{aligned}
				\end{equation}
				where $\eta$ is a factor to control variances.
				%
				Typically, we can use $T_S=100$ or $T_S=50$ for efficient synthesis.
				\footnotetext{The DDIM paper \cite{generative_ddim} uses the notation $\alpha_t$ to represent $\bar{\alpha}_t$, while we keep using $\bar{\alpha}_t$ to be consistent with the notations in the DDPM paper \cite{generative_ddpm}.}

			\textit{Conditional GAN.}
				GAN \cite{generative_gan} learns data synthesis by the adversarial learning between a generator $g_\theta(\cdot)$ and a discriminator $f_\omega(\cdot)$.
				In CGAN \cite{generative_cgan}, a generator takes noise $\bm{z}$ and conditions $\bm{c}$ as inputs to synthesize samples $g_\theta\left(\bm{z}, \bm{c}\right)$, 
				while a discriminator distinguishes real samples $\bm{x}$ from synthetic samples $g_\theta\left(\bm{z}, \bm{c}\right)$.
				Based on the Wasserstein loss \cite{generative_wgan}, the objective of CGAN can be formulated as:
				\begin{equation}
					\label{equ_cgan_train}
					\begin{aligned}
						\mathcal{L}\left(\theta, \omega\right) := \mathbb{E}_{\bm{x}} \left[f_\omega\left(\bm{x}, \bm{c}\right)\right] - \mathbb{E}_{\bm{z}} \left[f_\omega\left(g_\theta\left(\bm{z}, \bm{c}\right), \bm{c}\right)\right],
					\end{aligned}
				\end{equation}
				where $f_\omega\left(\bm{x}, \bm{c}\right)$ denotes the outputs of $f_\omega(\cdot)$ when its inputs are real samples, and $f_\omega\left(g_\theta\left(\bm{z}, \bm{c}\right), \bm{c}\right)$ denotes the outputs of $f_\omega(\cdot)$ when its inputs are synthetic samples.
				CGAN trains $f_\omega(\cdot)$ and $g_\theta\left(\cdot\right)$ alternately by $\min_\theta\max_\omega \mathcal{L}\left(\theta, \omega\right)$, where $f_\omega(\cdot)$ aims to maximize the distances between $\bm{x}$ and $g_\theta\left(\bm{z}, \bm{c}\right)$, while $g_\theta\left(\cdot\right)$ aims to minimize such distances.
				After training $f_\omega(\cdot)$ and $g_\theta\left(\cdot\right)$, we can use the generator to synthesize $\hat{\bm{x}}=g_\theta\left(\bm{z},\bm{c}\right)$.

			\noindent\textbf{Training Encoder-decoder Models.}
				After training a generative model, GenHPE exploits it to synthesize RF signals conditioned on counterfactual skeletons as Equations (\ref{equ_ske_mani_0})$\sim$(\ref{equ_ske_mani_2}).
				The synthetic RF signals are aggregated into $\bm{r}$, 
				which regularizes an encoder-decoder model for 3D HPE as Equations (\ref{equ_drl_1})$\sim$(\ref{equ_loss_sum}).
				We design an encoder with a simplified U-Net \cite{exp_unet} and construct a two-stream decoder with self-attention layers augmented by convolutional layers.
				%
				We only use data in source domains for training, while encoder-decoder models can generalize to target domains thanks to domain-independent representation learning with generative counterfactuals and difference-based aggregation.
				%
				More details about models are provided in \ul{Appendix \ref{appendix_encoder_decoder}}.
		
	%
	%
	%
	%
	%
	%
	%
	%
	%
	%
	%
	%
	%
	%
	%
	%
	%
	%
	%
	%
	\section{Experiments}
		\label{sec_exp}

		\subsection{Datasets}
			We evaluate GenHPE on three public datasets of different RF signal sources, including WiFi, UWB, and mmWave.
			(1) The \textbf{WiFi} dataset \cite{method_piw_3d} includes 7 subjects in 3 environments, where we extract 30707 CSI samples for evaluation. 
			%
			%
			(2) The \textbf{UWB} dataset \cite{data_operanet} collects about 8 hours of data from 6 subjects in 2 environments.
			As suggested by the authors, we segment the UWB data into 293930 samples.
			%
			%
			(3) The \textbf{mmWave} dataset \cite{data_mmfi} consists of over 320000 point cloud frames from 40 subjects in 4 environments.
			%
			%
			In line with the authors, we combine frames to obtain 95666 samples. 
			%
			
			%
			We apply three data split strategies to evaluate GenHPE in different scenarios.
			(1) \textbf{Random}: We shuffle each dataset and randomly split it into a training set (80\%), a validation set (10\%), and a test set (10\%).
			(2) \textbf{Cross-Subject}: For each dataset, we randomly select the data of certain subjects as a training set and the data of other subjects as a test set.
			(3) \textbf{Cross-Environment}: For each dataset, we randomly select the data from certain environments for training and the data from other environments for testing.
			Table \ref{table_data_split} shows the details of Cross-Subject/Environment splits,
			where we further use 10\% of each training set as a validation set.
			We train models on training sets and apply validation sets to select the best models for evaluation on test sets.
			%
			We further introduce these datasets and data splits in \ul{Appendix \ref{appendix_exp_dataset}}.
		
		%
		\subsection{Baselines}
			For 3D HPE with WiFi and UWB, whose signals are multichannel sequences, we compare GenHPE with four state-of-the-art methods.
			(1) \textbf{WiPose} \cite{method_wipose} combines CNNs and RNNs to learn spatial-temporal features.
			(2) \textbf{WPNet} \cite{method_wpnet} implements a ResNet \cite{related_resnet} to extract deep implicit features.
			(3) \textbf{WPFormer} \cite{method_wpformer} establishes Transformer layers for advanced RF-based 3D HPE.
			(4) \textbf{PiW-3D} \cite{method_piw_3d} further exploits a CSI encoder, a pose decoder, and a refine decoder to outperform other 3D HPE methods.

			For 3D HPE with mmWave, whose signals are point cloud data, we use four latest methods for comparison.
			(1) \textbf{Point Transformer} \cite{method_pt} supports 3D HPE from mmWave point clouds using self-attention layers.
			(2) \textbf{PoseFormer} \cite{method_poseformer} designs a spatial-temporal transformer to further strengthen 3D HPE.
			(3) \textbf{DiffPose} \cite{method_diffpose} formulates 3D HPE as a reverse diffusion process with GCNs as backbones, significantly reducing estimation errors.
			(4) \textbf{mmDiff} \cite{method_mmdiff} integrates diffusion models with global-local contexts to achieve better HPE accuracy than counterparts.

		\begin{table}[t]
			\footnotesize 
			\centering
			\begin{tabular}{lc@{\quad}cc@{\quad}cc@{\quad}c}
				
				\toprule
				
				\multirow{2}*{Data Split}		& \multicolumn{2}{c}{WiFi \cite{method_piw_3d}}	& \multicolumn{2}{c}{UWB \cite{data_operanet}}	& \multicolumn{2}{c}{mmWave \cite{data_mmfi}}	\\
				
				\cmidrule(lr){2-3}\cmidrule(lr){4-5}\cmidrule(lr){6-7}
				
				~					& Train		& Test	& Train		& Test 	& Train 	& Test	\\ 
				
				\midrule
				
				Cross-Subject 		& 6			& 1		& 5			& 1		& 32		& 8		\\ 		
				Cross-Environment 	& 2			& 1		& 1			& 1		& 3			& 1		\\
				
				\bottomrule
				
			\end{tabular}
			\caption{Numbers of subject(s)/environment(s) in training/test sets following previous works \cite{method_piw_3d, data_mmfi, related_widar3} for comparison.}
			\vspace{-1mm}
			\label{table_data_split}
		\end{table}

		\subsection{Evaluation Metrics}
			Following previous works \cite{method_piw_3d,method_mmdiff}, we apply three metrics for evaluation, including Mean Per Joint Position Error (\textbf{MPJPE}), Procrustes-aligned MPJPE (\textbf{PA-MPJPE}), and Mean Per Joint Dimension Location Error (\textbf{MPJDLE}). 
			We further describe these metrics in \ul{Appendix \ref{appendix_exp_metrics}}.

		\begin{table*}[t]
			\footnotesize 
			\centering
			\begin{tabular}{@{\ }l@{\ \ }c@{\ }c@{\ \ }c@{\ \ }c@{\ }c@{\ }c @{\ }c@{}c@{\ }c@{\ }}
				
				\toprule
				
				\multirow{2}*{Methods}	& \multicolumn{3}{c}{Random}	& \multicolumn{3}{c}{Cross-Subject}	& \multicolumn{3}{c}{Cross-Environment}	\\
				
				\cmidrule(lr){2-4}\cmidrule(lr){5-7}\cmidrule(lr){8-10}
				
				~	& MPJPE 	& PA-MPJPE 		& MPJDLE 	& MPJPE 	& PA-MPJPE 		& MPJDLE 	& MPJPE 	& PA-MPJPE 		& MPJDLE	\\ 
				
				\midrule
				\rowcolor[HTML]{EFEFEF} \multicolumn{10}{c}{WiFi \cite{method_piw_3d}}\vspace{0.8mm}\\
				WiPose \cite{method_wipose}		& 105.66\std{1.03\ \ }	& 49.52\std{0.53}		& 49.42\std{0.46}		& 274.03\std{32.30}			& 125.75\std{24.21}		& 129.63\std{15.47}			& \ \ 444.72\std{169.37}	& \ \ 203.36\std{59.98}	& 214.59\std{82.33}		\\
				WPNet \cite{method_wpnet}		& 98.24\std{1.97}		& 46.02\std{0.83}		& 45.60\std{0.88}		& \ti{259.74\std{34.87}}	& \ti{121.83\std{23.81}}& \ti{122.12\std{16.02}}	& \ti{239.48\std{5.10\ \ }}	& \ti{112.91\std{6.68}}	& \ti{112.23\std{2.48\ \ }}	\\
				WPFormer \cite{method_wpformer}	& 93.93\std{3.32}		& \ti{41.99\std{0.64}}	& 43.68\std{1.71}		& 272.74\std{33.79}			& 125.75\std{25.65}		& 127.72\std{15.98}			& 259.38\std{24.10}			& \ \ 122.80\std{11.68}	& 121.15\std{12.77}	\\
				PiW-3D \cite{method_piw_3d}		& \ti{91.78\std{2.03}}	& 43.00\std{1.01}		& \ti{42.44\std{0.94}}	& 268.50\std{32.90}			& 125.78\std{24.85}		& 126.17\std{15.79}			& 252.49\std{7.89\ \ }		& 123.47\std{7.77}		& 119.62\std{4.66\ \ }	\\
				GenHPE$_1$ (Ours)				& \tb{89.16\std{0.98}}	& \tb{41.40\std{0.36}}	& \tb{41.12\std{0.42}}	& 211.81\std{2.84\ \ }		& \tb{95.21\std{1.74}}	& 101.94\std{1.67\ \ }		& \tb{228.88\std{0.56\ \ }}	& \tb{105.81\std{0.90}}	& \tb{108.17\std{0.24\ \ }}	\\
				GenHPE$_2$ (Ours) 				& 89.42\std{1.11}		& 42.09\std{0.59}		& 41.39\std{0.53}		& \tb{211.68\std{2.69\ \ }}	& \tb{95.21\std{1.74}}	& \tb{101.88\std{1.59\ \ }}	& 228.88\std{0.57\ \ }		& \tb{105.81\std{0.90}}	& 108.17\std{0.25\ \ }	\\
				\midrule
				\rowcolor[HTML]{EFEFEF} \multicolumn{10}{c}{UWB \cite{data_operanet}}\vspace{0.8mm}\\ 	
				
				WiPose \cite{method_wipose}		& 108.42\std{1.21\ \ }	& 70.62\std{0.61}		& 51.40\std{0.56}		& 332.96\std{37.50}			& 147.10\std{8.19\ \ }		& 151.95\std{14.33}			& 316.42\std{25.58}			& 145.00\std{6.19}		& 145.41\std{10.49}	\\
				WPNet \cite{method_wpnet}		& 94.44\std{1.30}		& 60.92\std{0.75}		& 44.73\std{0.59}		& \ti{331.34\std{38.04}}	& \ti{144.02\std{6.31\ \ }}	& \ti{150.81\std{14.70}}	& \ti{303.34\std{28.19}}	& \ti{137.79\std{7.62}}	& \ti{139.46\std{11.67}}	\\
				WPFormer \cite{method_wpformer}	& 91.52\std{0.96}		& 59.09\std{0.52}		& 43.51\std{0.45}		& 335.79\std{38.01}			& 146.58\std{7.73\ \ }		& 153.12\std{14.44}			& 308.93\std{26.30}			& 141.53\std{6.96}		& 142.43\std{10.69}	\\
				PiW-3D \cite{method_piw_3d} 	& \ti{76.57\std{1.22}}	& \ti{52.98\std{0.65}}	& \ti{36.76\std{0.57}}	& 336.68\std{38.61}			& 147.85\std{8.24\ \ }		& 153.50\std{14.89}			& 310.67\std{27.24}			& 144.24\std{6.81}		& 143.43\std{11.20}	\\
				
				GenHPE$_1$ (Ours) 				& 75.59\std{1.02}		& \tb{51.43\std{0.47}}	& 36.14\std{0.47}		& \tb{279.17\std{3.99\ \ }}	& \tb{124.42\std{1.10\ \ }}	& \tb{128.79\std{1.46\ \ }}	& \tb{292.80\std{12.66}}	& 133.97\std{6.52}		& \tb{135.36\std{5.84\ \ }}	\\
				GenHPE$_2$ (Ours) 				& \tb{75.24\std{0.64}}	& 51.48\std{0.49}		& \tb{36.03\std{0.31}}	& 279.25\std{3.12\ \ }		& 125.28\std{1.03\ \ }		& 129.01\std{1.06\ \ }		& 302.44\std{20.76}			& \tb{133.44\std{7.72}}	& 139.23\std{8.34\ \ }	\\

				\midrule
				\rowcolor[HTML]{EFEFEF} \multicolumn{10}{c}{mmWave \cite{data_mmfi}}\vspace{0.8mm}\\
				Point Transformer \cite{method_pt}	& 106.47\std{2.50\ \ }	& 65.46\std{1.40}		& 52.33\std{1.29}		& 123.09\std{3.53\ \ }		& 71.71\std{2.14}		& 60.36\std{1.66}		& 130.39\std{10.89}			& \ \ 77.72\std{4.98}		& 63.95\std{5.40}	\\
				PoseFormer \cite{method_poseformer} & 99.96\std{1.42}		& 62.12\std{0.99}		& 49.03\std{0.70}		& 115.77\std{3.62\ \ }		& 67.56\std{1.49}		& 56.70\std{1.66}		& 121.43\std{14.36}			& \ \ 69.87\std{4.57}		& 59.57\std{7.19}	\\
				DiffPose \cite{method_diffpose} 	& 89.36\std{2.48}		& 53.37\std{1.69}		& 43.91\std{1.21}		& \ti{106.72\std{3.48\ \ }}	& 58.80\std{0.83}		& \ti{52.25\std{1.54}}	& \ti{113.04\std{11.98}}	& \ \ 62.28\std{4.57}		& \ti{55.42\std{5.78}}	\\
				mmDiff \cite{method_mmdiff}			& \ti{79.38\std{2.03}}	& \ti{48.22\std{1.28}}	& \ti{39.03\std{0.99}}	& 107.50\std{3.47\ \ }		& \ti{57.01\std{1.99}}	& 52.65\std{1.55}		& 113.20\std{14.19}			& \ \ \ti{60.06\std{5.72}}	& 55.47\std{6.97}	\\
				GenHPE$_1$ (Ours) 					& 78.36\std{1.95}		& 47.51\std{1.04}		& 38.44\std{0.96}		& 107.54\std{2.88\ \ }		& 57.93\std{1.47}		& 52.67\std{1.40}		& 112.12\std{2.21\ \ }		& \ \ 58.83\std{1.12}		& 54.56\std{1.07}	\\
				GenHPE$_2$ (Ours) 					& \tb{77.30\std{2.19}}	& \tb{45.84\std{0.82}}	& \tb{37.96\std{1.07}}	& \tb{105.83\std{2.09\ \ }}	& \tb{56.68\std{1.30}}	& \tb{51.89\std{1.05}}	& \tb{111.19\std{2.77\ \ }}	& \tb{\ \ 57.61\std{1.10}}	& \tb{54.13\std{1.35}}	\\
				%
				%
				%
				\bottomrule
				
			\end{tabular}
			\caption{Performance of 3D HPE methods in three scenarios with WiFi, UWB, and mmWave. GenHPE$_1$ adopts DDPM \cite{generative_ddpm} with 1000 synthesis steps.  GenHPE$_2$ adopts DDIM \cite{generative_ddim} with 100 synthesis steps. \textbf{Bold}: the best results. \ti{Blue}: the best baseline results. (Unit: mm)} 
			\vspace{-1mm}
			\label{table_result_main}
		\end{table*}

		\subsection{Implementation Details}
			%
			We explore DDPM \cite{generative_ddpm}, DDIM \cite{generative_ddim}, and CGAN \cite{generative_cgan} for counterfactual synthesis.
			To optimize DDPM, we adopt a linear spaced noise schedule where $\beta_t\in\left[10^{-5}, 10^{-1}\right]$ with diffusion steps $T=1000$.
			The learning rate of DDPM is $10^{-4}$.
			DDIM reuses the trained neural networks in DDPM with fewer synthesis steps.
			We employ DDIM with synthesis steps $T_S=100$ or $T_S=50$ and set the variance factor $\eta=0$.
			%
			To optimize CGAN, the generator is trained with the learning rate of $2\times10^{-4}$, and the discriminator is trained with the learning rate of $10^{-4}$.
			%
			%
			All generative models are trained for $1000$ epochs with the batch size of $128$.

			Both generative models and encoder-decoder models are optimized by Adam \cite{exp_adam} on a single Nvidia RTX 4090 GPU.
			We employ Mean Squared Error (MSE) as $\mathcal{L}_{\mathrm{pe}}$ and $\mathcal{L}_{\mathrm{cr}}$ with $\lambda=1.0$.
			%
			%
			We conduct a hyper-parameter sensitivity study to discuss different values of $\lambda$ in \ul{Appendix \ref{appendix_exp_sensitivity}}.
			We use the learning rate of $10^{-4}$ to optimize encoder-decoder models for $200$ epochs with the batch size of $32$.
			All experiments are based on PyTorch 2.0.1 with Python 3.9.12.
			We repeat each experiment $10$ times with random seeds to report the means and standard deviations of results.
			Note that each repetition randomly chooses subjects/environments for model evaluation with \textit{diverse test domains}.
		\subsection{Comparison with State-of-the-art Methods}
			%
			%
			GenHPE outperforms state-of-the-art methods with significantly decreased estimation errors in cross-subject and cross-environment scenarios, as shown in Table \ref{table_result_main}.

			In random split scenarios, we evaluate 3D HPE performance in known domains, which have been widely discussed \cite{method_piw_3d,method_mmdiff}.
			GenHPE outperforms the best baselines with slightly lower estimation errors.
			This is reasonable because in random split scenarios, existing methods can directly learn the confounder distributions of all domains. 
			Thus, the effectiveness of confounder elimination in GenHPE seems less notable.
			%
			However, GenHPE still makes contributions to the generalization ability of 3D HPE.
			
			%
			In cross-subject scenarios, we can evaluate how GenHPE eliminates subject-specific confounders and generalizes to new subjects.
			In contrast to the best baselines, GenHPE improves WiFi-based 3D HPE with 48.06mm lower MPJPE, 26.62mm lower PA-MPJPE, and 20.24mm lower MPJDLE.
			With UWB, GenHPE reduces three metrics by 52.17mm, 19.60mm, and 22.02mm, respectively, compared to the best baselines.
			We can observe the significantly lower standard deviations of GenHPE ($\leq$2.69mm with WiFi and $\leq$3.99mm with UWB) against those of baselines ($\leq$34.87mm with WiFi and $\leq$38.04mm with UWB). 
			The improvement of GenHPE with mmWave is less obvious, since point cloud signals from high-frequency mmWave contain less confounders, but GenHPE still outperforms all baselines.
			These results illustrate that GenHPE indeed eliminates subject-specific confounders for more robust cross-subject 3D HPE.
			
			%
			In cross-environment scenarios, we can validate how GenHPE eliminates environment-specific confounders and generalizes to new environments.
			WiFi-based 3D HPE benefits from GenHPE with 10.60mm lower MPJPE, 7.10mm lower PA-MPJPE, and 4.06mm lower MPJDLE, against the best baselines.
			%
			With UWB, GenHPE outperforms the best baselines by 10.54mm on MPJPE, 4.35mm on PA-MPJPE, and 4.1mm on MPJDLE.
			%
			Similar to cross-subject scenarios, GenHPE also results in obviously lower standard deviations than baselines. 
			GenHPE also achieves state-of-the-art performance for mmWave-based 3D HPE.
			%
			These results prove that GenHPE can effectively tackle environment-specific confounders for more robust cross-environment 3D HPE.

		\begin{table*}[t]
			\footnotesize 
			\centering
			\begin{tabular}{@{\ }l@{\ \ }c@{\ }c@{\ \ }c@{\ \ }c@{\ }c@{}c@{\ }c@{\ }c@{\ }c@{\ }}
				
				\toprule
				
				\multirow{2}*{Methods}	& \multicolumn{3}{c}{Random}	& \multicolumn{3}{c}{Cross-Subject}	& \multicolumn{3}{c}{Cross-Environment}	\\
				
				\cmidrule(lr){2-4}\cmidrule(lr){5-7}\cmidrule(lr){8-10}
				
				~	& MPJPE 	& PA-MPJPE 		& MPJDLE 	& MPJPE 	& PA-MPJPE 		& MPJDLE 	& MPJPE 	& PA-MPJPE 		& MPJDLE	\\ 
				
				\midrule
				\rowcolor[HTML]{EFEFEF} \multicolumn{10}{c}{WiFi \cite{method_piw_3d}}\vspace{0.8mm}\\
				(1) w/ DDPM ($T$ = 1000) 		& \tb{89.16\std{0.98}}	& \tb{41.40\std{0.36}}	& \tb{41.12\std{0.42}}	& 211.81\std{2.84}	& \tb{95.21\std{1.74}}	& 101.94\std{1.67}	& 228.88\std{0.56}	& 105.81\std{0.90}	& \tb{108.17\std{0.24}}	\\
				(2) w/ DDIM ($T_S$ = 100) 		& 89.42\std{1.11}	& 42.09\std{0.59}	& 41.39\std{0.53}	& 211.68\std{2.69}	& \tb{95.21\std{1.74}}	& 101.88\std{1.59}	& 228.88\std{0.57}	& 105.81\std{0.90}	& 108.17\std{0.25}	\\
				(3) w/ DDIM ($T_S$ = 50) 		& 89.51\std{1.30}	& 42.10\std{0.63}	& 41.43\std{0.58}	& 211.69\std{2.79}	& \tb{95.21\std{1.74}}	& 101.89\std{1.65}	& 228.88\std{0.57}	& 105.81\std{0.90}	& 108.17\std{0.25}	\\
				(4) w/ CGAN 			& 91.28\std{1.48}	& 43.07\std{0.61}	& 42.30\std{0.69}	& \tb{211.46\std{2.71}}	& \tb{95.21\std{1.74}}	& \tb{101.78\std{1.60}}	& \tb{228.87\std{0.56}}	& \tb{105.80\std{0.90}}	& 108.17\std{0.25}	\\
				(5) w/o Skeleton Embedding 	& 90.70\std{1.50}	& 42.49\std{0.54}	& 41.93\std{0.72}	& 211.83\std{2.81}	& 95.21\std{1.75}	& 101.96\std{1.67}	& \ \ 258.56\std{10.07}	& 117.98\std{7.15}	& 123.68\std{6.01}	\\
				(6) w/o Gen. Counterfactuals& 93.26\std{0.73}	& 44.48\std{0.32}	& 43.37\std{0.34}	& 260.59\std{4.43}	& 112.64\std{1.13\ \ }	& 122.62\std{2.15}	& \ \ 261.86\std{12.97}	& 117.22\std{7.80}	& 126.25\std{8.01}	\\
				(7) w/ Decoder Only 			& 122.31\std{2.02\ \ }	& 56.46\std{0.86}	& 56.73\std{0.93}	& 255.63\std{5.45}	& 113.55\std{1.36\ \ }	& 120.07\std{2.35}	& 254.77\std{6.77}	& 126.18\std{7.08}	& 120.39\std{4.30}	\\
				\midrule
				\rowcolor[HTML]{EFEFEF} \multicolumn{10}{c}{UWB \cite{data_operanet}}\vspace{0.8mm}\\ 	
				
				(1) w/ DDPM ($T$ = 1000)		& 75.59\std{1.02}	& 51.43\std{0.47}	& 36.14\std{0.47}	& 279.17\std{3.99}	& \tb{124.42\std{1.10\ \ }}	& 128.79\std{1.46}	& \ \ 292.80\std{12.66}	& 133.97\std{6.52}	& 135.36\std{5.84}	\\
				(2) w/ DDIM ($T_S$ = 100) 		& 75.24\std{0.64}	& 51.48\std{0.49}	& 36.03\std{0.31}	& 279.25\std{3.12}	& 125.28\std{1.03\ \ }	& 129.01\std{1.06}	& \ \ 302.44\std{20.76}	& \tb{133.44\std{7.72}}	& 139.23\std{8.34}	\\
				(3) w/ DDIM ($T_S$ = 50) 		& \tb{74.97\std{0.49}}	& \tb{51.18\std{0.42}}	& \tb{35.91\std{0.26}}	& \tb{277.45\std{1.86}}	& 125.02\std{1.01\ \ }	& \tb{128.34\std{0.79}}	& \ \ 294.95\std{10.53}	& 134.60\std{4.24}	& 136.04\std{4.47}	\\
				(4) w/ CGAN 			& 77.60\std{0.81}	& 52.69\std{0.54}	& 37.12\std{0.37}	& 277.83\std{4.19}	& 125.11\std{1.18\ \ }	& 128.51\std{1.56}	& 294.41\std{7.52}	& 134.35\std{3.69}	& 136.05\std{3.65}	\\
				(5) w/o Skeleton Embedding 	& 77.37\std{0.79}	& 52.57\std{0.41}	& 36.97\std{0.37}	& 279.48\std{3.98}	& 125.58\std{0.92\ \ }	& 129.08\std{1.41}	& \tb{\ \ 291.14\std{10.54}}	& 134.25\std{5.99}	& \tb{134.65\std{5.02}}	\\
				(6) w/o Gen. Counterfactuals& 83.34\std{0.31}	& 57.09\std{0.19}	& 39.93\std{0.16}	& 298.99\std{2.61}	& 135.73\std{0.45\ \ }	& 138.81\std{1.00}	& 337.45\std{3.45}	& 146.74\std{0.49}	& 153.21\std{1.14}	\\
				(7) w/ Decoder Only 			& 97.31\std{0.94}	& 65.79\std{0.54}	& 46.56\std{0.44}	& 303.16\std{2.55}	& 136.35\std{0.60\ \ }	& 139.86\std{0.94}	& 332.82\std{2.70}	& 146.31\std{0.36}	& 151.46\std{1.02}\\
				\midrule
				\rowcolor[HTML]{EFEFEF} \multicolumn{10}{c}{mmWave \cite{data_mmfi}}\vspace{0.8mm}\\
				(1) w/ DDPM ($T$ = 1000) & 78.36\std{1.95}	& 47.51\std{1.04}	& 38.44\std{0.96}	& 107.54\std{2.88}	& 57.93\std{1.47}	& \ \ 52.67\std{1.40}	& 112.12\std{2.21}	& \ \ 58.83\std{1.12}	& \ \ 54.56\std{1.07}	\\
				(2) w/ DDIM ($T_S$ = 100) & \tb{77.30\std{2.19}}	& \tb{45.84\std{0.82}}	& \tb{37.96\std{1.07}}	& 105.83\std{2.09}	& 56.68\std{1.30}	& \tb{\ \ 51.89\std{1.05}}	& \tb{111.19\std{2.77}}	& \tb{\ \ 57.61\std{1.10}}	& \tb{\ \ 54.13\std{1.35}}	\\
				(3) w/ DDIM ($T_S$ = 50) 	& 77.70\std{1.45}	& 46.14\std{0.73}	& 38.15\std{0.71}	& \tb{105.82\std{2.00}}	& \tb{55.96\std{1.01}}	& \ \ 51.90\std{0.99}	& 112.26\std{2.11}	& \ \ 58.50\std{1.78}	& \ \ 54.68\std{0.98}	\\
				(4) w/ CGAN 		& 79.10\std{2.24}	& 47.70\std{1.27}	& 38.82\std{1.07}	& 107.49\std{1.54}	& 57.44\std{0.76}	& \ \ 52.68\std{0.73}	& 114.56\std{2.25}	& \ \ 60.51\std{1.30}	& \ \ 55.73\std{1.09}	\\
				(5) w/o Skeleton Embedding 	& 78.17\std{2.28}	& 46.22\std{0.94}	& 38.37\std{1.12}	& 109.83\std{2.21}	& 60.46\std{1.33}	& \ \ 53.82\std{1.12}	& 112.61\std{1.63}	& \ \ 59.44\std{1.21}	& \ \ 54.81\std{0.80}	\\
				(6) w/o Gen. Counterfactuals & 84.17\std{1.37}	& 50.11\std{0.82}	& 41.34\std{0.66}	& 109.44\std{1.49}	& 59.76\std{1.17}	& \ \ 53.69\std{0.67}	& 113.43\std{3.49}	& \ \ 60.38\std{1.53}	& \ \ 55.30\std{1.69}	\\
				(7) w/ Decoder Only 			& 86.05\std{2.57}	& 51.32\std{1.11}	& 42.28\std{1.27}	& 109.11\std{1.97}	& 60.49\std{1.03}	& \ \ 53.55\std{1.01}	& 113.88\std{2.15}	& \ \ 61.00\std{0.74}	& \ \ 55.54\std{1.10}	\\
				\bottomrule
				
			\end{tabular}
			\caption{Ablation study of GenHPE with (w/) or without (w/o) different components. \textbf{Bold} highlights the best results.  (Unit: mm)}
			\vspace{-1mm}
			\label{table_result_ablation}
		\end{table*}

		\subsection{Ablation Study}
			%
			Table \ref{table_result_ablation} presents the ablation study of GenHPE to discuss the contributions of different components in our method. 

			\textit{Generative Models.}
			Generally, GenHPE achieves similar performance with different generative models.
			In most cases, using diffusion models leads to slightly better performance than using CGANs.
			Compared with DDPMs, using DDIMs not only enables effective counterfactual synthesis for GenHPE but also requires less synthesis steps as few as $T_S$ = 50.
			Such results demonstrate that most generative models are applicable in our method, and GenHPE does not rely on specific models to achieve state-of-the-art 3D HPE.
			
			%
			\begin{figure}[t]
				\centering
				\begin{subfigure}{0.49\linewidth}
					\centering
					\includegraphics[width=\linewidth]{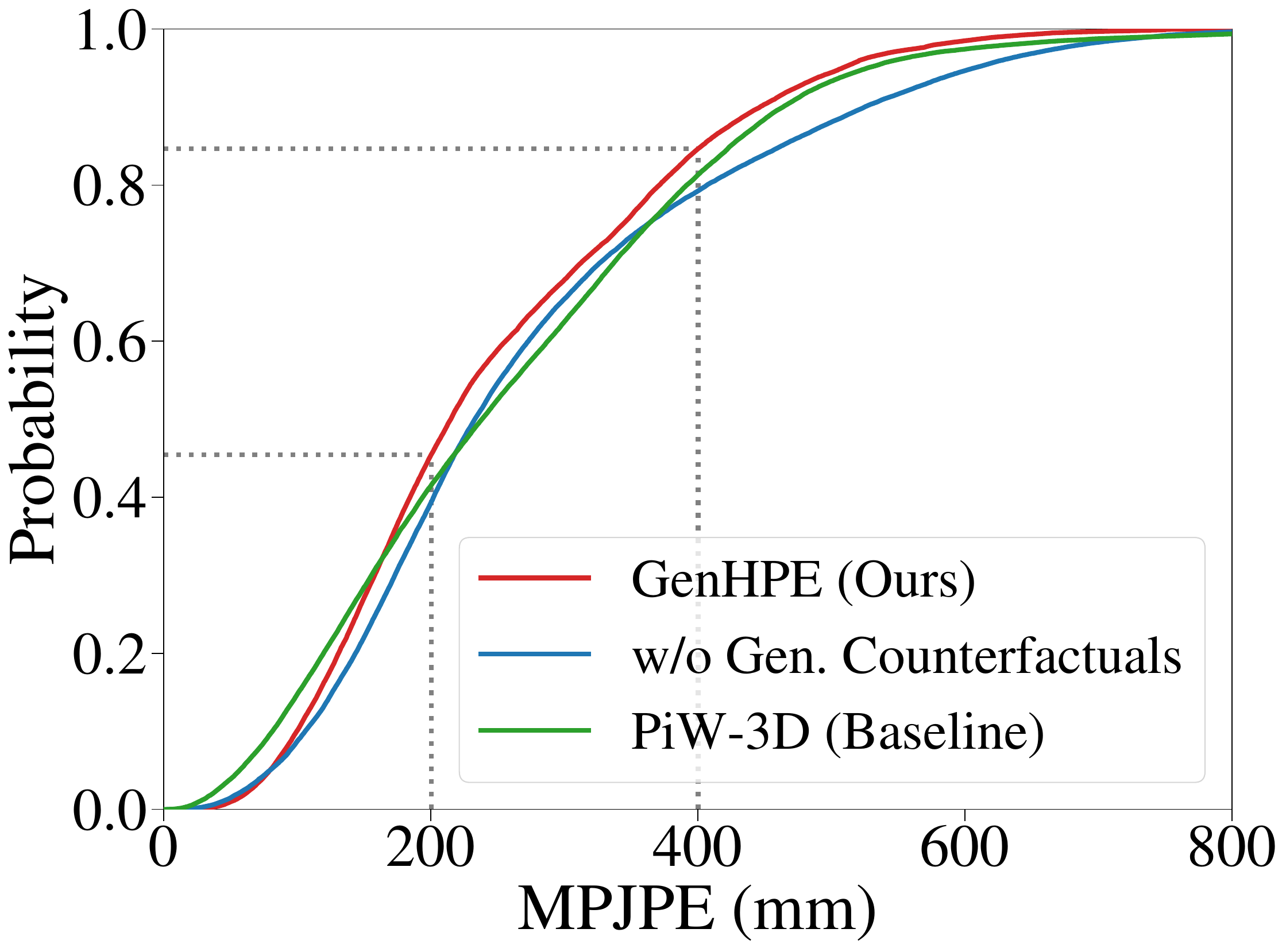}
					\caption{Cross-Subject.}
					\label{figure_cdf_cross_subject}
				\end{subfigure}
				\begin{subfigure}{0.49\linewidth}
					\centering
					\includegraphics[width=\linewidth]{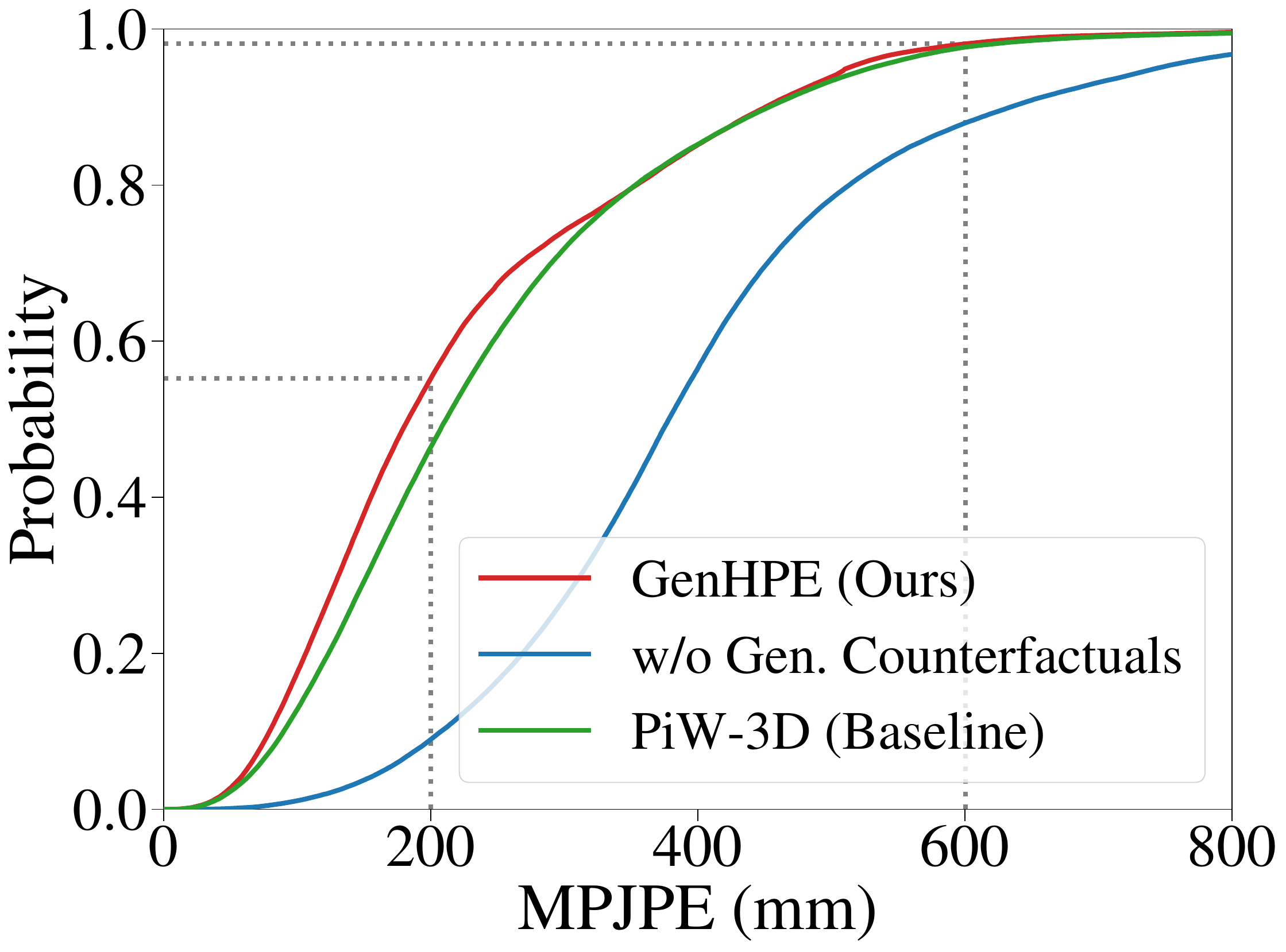}
					\caption{Cross-Environment.}
					\label{figure_cdf_cross_env}
				\end{subfigure}
				\caption{%
					Cumulative distributions of MPJPE (mm) for cross-subject/environment 3D HPE with WiFi \cite{method_piw_3d}.
				}
				\label{figure_cdf}
			\end{figure}

			\begin{figure}[t]
				\centering
				\begin{subfigure}{0.5\linewidth}
					\centering
					\includegraphics[width=\linewidth]{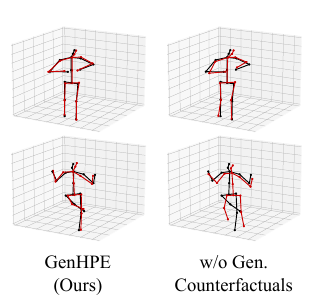}
					\caption{Cross-Subject.}
					\label{figure_mmfi_cross_subject}
				\end{subfigure}%
				\hfill
				\begin{subfigure}{0.5\linewidth}
					\centering
					\includegraphics[width=\linewidth]{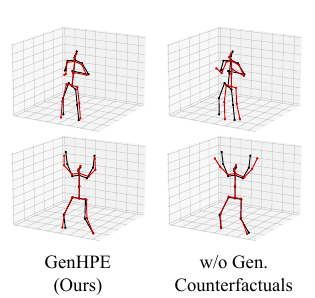}
					\caption{Cross-Environment.}
					\label{figure_mmfi_cross_env}
				\end{subfigure}
				\caption{%
					Performance of GenHPE with/without generative counterfactuals for cross-subject/environment 3D HPE with mmWave \cite{data_mmfi}.
					Black: ground-truth human poses. \textcolor[HTML]{CC0000}{Red}: model estimations.
				}
				\vspace{-8mm}
				\label{figure_example}
			\end{figure}

			\textit{Skeleton Embedding.}
			We remove Skeleton Embedding from GenHPE and directly use ground-truth/counterfactual human body joints as conditions for DDPMs with 1000 synthesis steps.
			After such ablation, GenHPE results in higher estimation errors in most cases, which supports our hypothesis that what really interfere with RF signals are different human body parts but not discrete human body joints.

			\textit{Generative (Gen.) Counterfactuals.}
			We directly examine encoder-decoder models for 3D HPE without generative counterfactuals.
			Such ablation leads to severe performance declines, especially in cross-subject and cross-environment scenarios.
			Figure \ref{figure_cdf} compares the cumulative distributions of MPJPE with WiFi to highlight the effectiveness of generative counterfactuals.
			Figure \ref{figure_example} further visualizes the performance of GenHPE with/without generative counterfactuals on the mmWave dataset.
			We further investigate the performance of using decoders only and observe even worse 3D HPE performance.
			%
			These results demonstrate that the regularization of generative counterfactuals indeed helps GenHPE generalize to new subjects and environments.
			
		%
		%
		%
		%
		%
		%
		%
		%
		%
		%
		%
		%
		%
		%
		%
		%
		%
		%
		%
		%
		\section{Conclusion}
			\label{sec_conclusion}
			GenHPE pioneers RF-based cross-domain 3D HPE by learning how human body parts and confounders interfere with RF signals.
			%
			%
			GenHPE synthesizes counterfactual RF signals conditioned on manipulated skeleton labels which ``remove'' human body parts.
			%
			A difference-based scheme eliminates domain-specific confounders using counterfactual signals, by which GenHPE regularizes an encoder-decoder model to learn domain-independent representations.
			%
		 	Such representations help GenHPE generalize to new subjects/environments, thereby outperforming state-of-the-art methods in cross-subject/environment scenarios.
			%
		
	\newpage
	\setcounter{table}{0}
	\setcounter{figure}{0}
	\renewcommand{\tablename}{Supplement Table}
	\renewcommand{\figurename}{Supplement Figure}
	\maketitlesupplementary
	\appendix
	%
%
%
%
%
%
%
%
%
%
%
%
%
%
%
%
%
%
%
%
\section{Models in GenHPE}
	GenHPE explores three generative models to synthesize counterfactual RF signals and implements an encoder-decoder model for 3D HPE.
	Herein, we provide more details about these models.

	\subsection{Generative Models}
		\label{appendix_gen_model}
		GenHPE leverages three generative models, including conditional DDPM \cite{generative_ddpm,generative_cdm}, DDIM \cite{generative_ddim}, and CGAN \cite{generative_cgan,generative_wgan}.
		\subsubsection{Conditional DDPM}
			DDPM \cite{generative_ddpm} aims to approximate a data distribution $q\left(\bm{x}_0\right)$ by learning a model distribution $p_\theta\left({\bm{x}}_0\right)$.
			Specifically, DDPM introduces a fixed forward process and a learnable reverse process.
			The forward process converts $q\left(\bm{x}_0\right)$ to a Gaussian distribution $q\left(\bm{x}_T\right)=\mathcal{N}\left(\bm{x}_T;\textbf{0}, \textbf{I}\right)$ in $T$ steps.
			The reverse process learns to convert a Gaussian distribution $p\left(\hat{\bm{x}}_T\right)=\mathcal{N}\left(\hat{\bm{x}}_T;\textbf{0}, \textbf{I}\right)$ to the model distribution $p_\theta\left(\hat{\bm{x}}_0\right)$ in $T$ steps.
			The objective of DDPMs is to maximize the likelihood function $\mathbb{E}\left[p_\theta\left(\bm{x}_0\right)\right]$.

			The forward process is a fixed Markov chain that gradually adds noise to $\bm{x}_0\sim q\left(\bm{x}_0\right)$ with an increasing variance schedule $\left[\beta_1, ..., \beta_T\right]$.
			Each step in the forward process is a fixed Gaussian transition $q(\bm{x}_t|\bm{x}_{t-1})$ as:
			\begin{equation}
				\label{equ_ddpm_forward_step}
				\begin{aligned}
					q(\bm{x}_t|\bm{x}_{t-1}) = \mathcal{N} \left(\bm{x}_{t-1} ; \sqrt{1-\beta_t} \bm{x}_{t-1}, \beta_t \textbf{I}\right),
				\end{aligned}
			\end{equation}
			where $\beta_t\in\left(0,1\right)$ is the variance.
			Accordingly, the forward process can be formulated as:
			\begin{equation}
				\label{equ_ddpm_forward}
				\begin{aligned}
					q\left(\bm{x}_{1:T}|\bm{x}_0\right):= \prod_{t=1}^{T} q(\bm{x}_t|\bm{x}_{t-1}).
				\end{aligned}
			\end{equation}

			The reverse process aims to synthesize samples $\hat{\bm{x}}_0$ from Gaussian noise $\hat{\bm{x}}_T\sim \mathcal{N}\left(\hat{\bm{x}}_T;\textbf{0}, \textbf{I}\right)$ with a learnable $T$-step Markov chain as:
			\begin{equation}
				\label{equ_ddpm_reverse}
				p_\theta\left(\hat{\bm{x}}_{0:T}\right) := p\left(\hat{\bm{x}}_T\right) \prod_{t=1}^{T} p_\theta\left(\hat{\bm{x}}_{t-1}|\hat{\bm{x}}_t\right),
			\end{equation}
			where $p_\theta\left(\hat{\bm{x}}_{t-1}|\hat{\bm{x}}_t\right)$ is a learnable Gaussian transition as:
			\begin{equation}
				\label{equ_ddpm_reverse_step}
				p_\theta\left(\hat{\bm{x}}_{t-1}|\hat{\bm{x}}_t\right) := \mathcal{N} \left(\hat{\bm{x}}_{t-1} ; \bm{\mu}_\theta\left(\hat{\bm{x}}_t, t\right), \bm{\Sigma}_\theta\left(\hat{\bm{x}}_t, t\right) \right).
			\end{equation}
			$\theta$ denotes the learnable parameters of Gaussian transition with the mean $\bm{\mu}_\theta\left(\hat{\bm{x}}_t, t\right)$ and the variance $\bm{\Sigma}_\theta\left(\hat{\bm{x}}_t, t\right)$.

			To maximize the likelihood $\mathbb{E}\left[p_\theta\left(\bm{x}_0\right)\right]$, it is equivalent to minimize the negative log likelihood $\mathbb{E}\left[-\log p_\theta\left(\bm{x}_0\right)\right]$.
			Given a variational upper bound on $\mathbb{E}\left[-\log p_\theta\left(\bm{x}_0\right)\right]$, to minimize the negative log likelihood is equivalent to minimize the Kullback-Leibler (KL) divergence $D_{\mathrm{KL}}$ as: 
			\begin{equation}
				\label{equ_ddpm_kl}
				\min_\theta D_{\mathrm{KL}}\left(q\left(\bm{x}_{t-1}|\bm{x}_{t},\bm{x}_{0}\right) \|\ p_\theta\left(\hat{\bm{x}}_{t-1}|\hat{\bm{x}}_{t}\right)\right),
			\end{equation}
			between $p_\theta\left(\hat{\bm{x}}_{t-1}|\hat{\bm{x}}_{t}\right)$ and the forward process posterior $q\left(\bm{x}_{t-1}|\bm{x}_{t},\bm{x}_{0}\right)$.
			Using the notations $\alpha_t:=1-\beta_t$ and $\bar{\alpha}_t:=\prod_{s=1}^{t}\alpha_s$, $q\left(\bm{x}_{t-1}|\bm{x}_{t},\bm{x}_{0}\right)$ can be presented as a fixed Gaussian distribution as:
			\begin{equation}
				\label{equ_ddpm_q0}
				\begin{aligned}
					q\left(\bm{x}_{t-1}|\bm{x}_{t},\bm{x}_{0}\right) =&\ \mathcal{N} \left(\bm{x}_{t-1}; \tilde{\bm{\mu}}_t, \tilde{\beta}_t\textbf{I}\right)	\\
					\mathrm{where}\quad \tilde{\bm{\mu}}_t :=& \frac{\sqrt{\bar{\alpha}_{t-1}}\beta_t}{1-\bar{\alpha}_t}\bm{x}_0 + \frac{\sqrt{\alpha_t}\left(1-\bar{\alpha}_{t-1}\right)}{1-\bar{\alpha}_{t}}	\bm{x}_t\\
					\mathrm{and}\quad \tilde{\beta}_t :=& \frac{1-\bar{\alpha}_{t-1}}{1-\bar{\alpha}_{t}}\beta_t.
				\end{aligned}
			\end{equation}
			Since the forward process is a fixed Markov chain, we have:
			\begin{equation}
				\label{equ_ddpm_q1}
				q\left(\bm{x}_{t}|\bm{x}_{0}\right) = \mathcal{N}\left(\bm{x}_t; \sqrt{\bar{\alpha}_t}\bm{x}_0, \left(1-\bar{\alpha}_t\right)\textbf{I}\right),
			\end{equation}
			which can be parameterized as $\bm{x}_t = \sqrt{\bar{\alpha}_t}\bm{x}_0 + \sqrt{1-\bar{\alpha}_t}\bm{\epsilon}$, where $\bm{\epsilon}\sim\mathcal{N}\left(\textbf{0}, \textbf{I}\right)$ denotes the added noise in each step.
			Accordingly, we can replace $\bm{x}_0$ with $\bm{x}_t$ in Equation (\ref{equ_ddpm_q0}) to derive $\tilde{\bm{\mu}}_t$ as:
			\begin{equation}
				\label{equ_ddpm_mu}
				\tilde{\bm{\mu}}_t = \frac{1}{\sqrt{\alpha_t}}\left(\bm{x}_t - \frac{\beta_t}{\sqrt{1-\bar{\alpha}_t}}\bm{\epsilon}\right).
			\end{equation}
			Given $\bm{x}_T\sim\mathcal{N}\left(\bm{x}_T;\textbf{0}, \textbf{I}\right)$, $\tilde{\bm{\mu}}_t$ and $\tilde{\beta}_t$ indicate how to reverse the forward process by Gaussian transitions, but $\bm{\epsilon}$ added in each step is unknown.
			To minimize the KL divergence between $q\left(\bm{x}_{t-1}|\bm{x}_{t},\bm{x}_{0}\right)$ and $p_\theta\left(\hat{\bm{x}}_{t-1}|\hat{\bm{x}}_t\right)$, the reverse process learns Gaussian transitions $p_\theta\left(\hat{\bm{x}}_{t-1}|\hat{\bm{x}}_t\right)$ with means $\bm{\mu}_\theta\left(\hat{\bm{x}}_t, t\right)$ and variances $\bm{\Sigma}_\theta\left(\hat{\bm{x}}_t, t\right)$ to fit $\tilde{\bm{\mu}}_t$ and $\tilde{\beta}_t$.
			Specifically, $\bm{\mu}_\theta\left(\hat{\bm{x}}_t, t\right)$ and $\bm{\Sigma}_\theta\left(\hat{\bm{x}}_t, t\right)$ are parameterized as:
			\begin{equation}
				\label{equ_ddpm_mu_sigma}
				\begin{aligned}
					& \bm{\mu}_\theta\left(\hat{\bm{x}}_t, t\right) = \frac{1}{\sqrt{\alpha_t}} \left( \hat{\bm{x}}_t - \frac{\beta_t}{\sqrt{1-\bar{\alpha}_t}} \bm{\epsilon}_\theta \left(\hat{\bm{x}}_t, t\right) \right), \\
					& \bm{\Sigma}_\theta\left(\hat{\bm{x}}_t, t\right) =\sigma_t^2\textbf{I}\ \ \mathrm{where}\ \ \sigma_t^2 = \frac{1-\bar{\alpha}_{t-1}}{1-\bar{\alpha}_t} \beta_t.
				\end{aligned}
			\end{equation}
			To fit $\tilde{\bm{\mu}}_t$ with $\bm{\mu}_\theta\left(\hat{\bm{x}}_t, t\right)$, it is equivalent to learn a noise model $\bm{\epsilon}_\theta \left(\hat{\bm{x}}_t, t\right)$ to predict $\bm{\epsilon}$ from $\hat{\bm{x}}_t$. 
			For the training of $\bm{\epsilon}_\theta \left(\hat{\bm{x}}_t, t\right)$, DDPM uses $\hat{\bm{x}}_t = \bm{x}_t$ and formulates the loss function as:
			\begin{equation}
				\begin{aligned}
					\label{equ_ddpm_loss}
					\hspace{-2mm}\mathcal{L}\left(\theta\right):=&\mathbb{E}_{{\bm{x}}_0}\left[ \left \|  \bm{\epsilon} - \bm{\epsilon}_\theta\left( {\bm{x}}_t, t \right) \right \|^2 \right] \\
					=& \mathbb{E}_{{\bm{x}}_0}\left[\left \|  \bm{\epsilon} - \bm{\epsilon}_\theta\left( \sqrt{\bar{\alpha}_t} {\bm{x}}_0 + \sqrt{1-\bar{\alpha}_t} \bm{\epsilon}, t \right) \right \|^2 \right],
				\end{aligned}
			\end{equation}
			which can be optimized by gradient descent.
			After training $\bm{\mu}_\theta\left(\hat{\bm{x}}_t, t\right)$, it can be used to synthesize $\hat{\bm{x}}_0$ from $\hat{\bm{x}}_T\sim \mathcal{N}\left(\textbf{0}, \textbf{I}\right)$ by iterating $t\in\left[T,...,1\right]$ as:
			\begin{equation}
				\label{equ_ddpm_synthesis}
				\begin{aligned}
					& \hat{\bm{x}}_{t-1} =\frac{1}{\sqrt{\alpha_t}} \left( \hat{\bm{x}}_t - \frac{\beta_t}{\sqrt{1-\bar{\alpha}_t}} \bm{\epsilon}_\theta \left(\hat{\bm{x}}_t, t\right) \right) + \sigma_t\bm{z} \\
					& \mathrm{and}\ \ \sigma_t =\sqrt{\frac{1-\bar{\alpha}_{t-1}}{1-\bar{\alpha}_t}} \sqrt{\beta_t},
				\end{aligned}
			\end{equation}
			where $\bm{z}\sim\mathcal{N}\left(\textbf{0}, \textbf{I}\right)$ when $t>1$ and $\bm{z}=0$ when $t=1$.

			Original DDPM is unconditional, while recent works \cite{generative_cdm} have further introduced conditions to control the target outputs of DDPM.
			Given input conditions $\bm{c}$, conditional DDPM formulates the loss function as:
			\begin{equation}
				\begin{aligned}
					\label{equ_conditional_ddpm_loss}
					\hspace{-2mm}\mathcal{L}\left(\theta\right):=&\mathbb{E}_{{\bm{x}}_0}\left[ \left \|  \bm{\epsilon} - \bm{\epsilon}_\theta\left( {\bm{x}}_t, t, \bm{c} \right) \right \|^2 \right] \\
					=& \mathbb{E}_{{\bm{x}}_0}\left[\left \|  \bm{\epsilon} - \bm{\epsilon}_\theta\left( \sqrt{\bar{\alpha}_t} {\bm{x}}_0 + \sqrt{1-\bar{\alpha}_t} \bm{\epsilon}, t, \bm{c} \right) \right \|^2 \right],\hspace{-2mm}
				\end{aligned}
			\end{equation}
			After training $\bm{\epsilon}_\theta\left( \cdot \right)$, a sample $\hat{\bm{x}}_0$ conditioned on $\bm{c}$ can be synthesized from $\hat{\bm{x}}_T\sim \mathcal{N}\left(\textbf{0}, \textbf{I}\right)$ with $t\in\left[T,...,1\right]$ as:
			\begin{equation}
				\label{equ_conditional_ddpm_synthesis}
				\begin{aligned}
					\hat{\bm{x}}_{t-1} =\frac{1}{\sqrt{\alpha_t}} \left( \hat{\bm{x}}_t - \frac{\beta_t}{\sqrt{1-\bar{\alpha}_t}} \bm{\epsilon}_\theta \left(\hat{\bm{x}}_t, t, \bm{c}\right) \right) + \sigma_t\bm{z}.
				\end{aligned}
			\end{equation}

			In GenHPE, we apply skeleton embeddings $\bm{c}=\varphi\left(\bm{h}\right)$ as conditions, while $\bm{h}$ can be ground-truth or manipulated skeleton vectors.
			Algorithm \ref{algorithm_1} illustrates the training of conditional DDPM in GenHPE, and Algorithm \ref{algorithm_2} illustrates the synthesis with conditional DDPM in GenHPE.

			\begin{algorithm}[t]
				\caption{Training Conditional DDPM \cite{generative_ddpm}}
				\label{algorithm_1}
				\textbf{repeat}
				\begin{algorithmic}[1] 
					\STATE sample ground-truth 3D HPE labels $\bm{y}$
					\STATE obtain ground-truth RF signal samples $\bm{x} \sim q\left(\bm{x}|\bm{y}\right)$
					\STATE convert $\bm{y}$ to skeleton vectors $\bm{h}$
					\STATE calculate skeleton embeddings $\bm{c} = \varphi\left(\bm{h}\right)$
					\STATE sample $t \sim \mathrm{Uniform}\left( \{ 1, ..., T \} \right)$
					\STATE sample $\bm{\epsilon} \sim \mathcal{N}\left(\textbf{0}, \textbf{I}\right)$	
					\STATE take gradient descent steps on \\
					$\qquad\triangledown_\theta \left \| \bm{\epsilon} - \bm{\epsilon}_\theta\left( \sqrt{\bar{\alpha}_t} \bm{x}_0 + \sqrt{1-\bar{\alpha}_t} \bm{\epsilon}, t, \bm{c} \right) \right \| ^2 $
				\end{algorithmic}
				\textbf{until} converged
			\end{algorithm}

			\begin{algorithm}[t]
				\caption{Synthesis with Conditional DDPM \cite{generative_ddpm}}
				\label{algorithm_2}
				\textbf{Input:} ground-truth or manipulated skeleton vectors $\bm{h}$ 
				
				\begin{algorithmic}[1] 
					\STATE sample $\hat{\bm{x}}_T \sim \mathcal{N}\left(\textbf{0}, \textbf{I}\right)$
					\STATE calculate skeleton embeddings $\bm{c} = \varphi\left(\bm{h}\right)$
					\FOR {$t = T, ..., 1$}
					\STATE $\bm{z} \sim \mathcal{N}\left(\textbf{0}, \textbf{I}\right)$\ \ if $t>1$\ \ else $\bm{z}=0$	
					\STATE $\sigma_t =\sqrt{\frac{1-\bar{\alpha}_{t-1}}{1-\bar{\alpha}_t}} \sqrt{\beta_t}$
					\STATE $\hat{\bm{x}}_{t-1} = \frac{1}{\sqrt{\alpha_t}} \left( \hat{\bm{x}}_t - \frac{\beta_t}{\sqrt{1-\bar{\alpha}_t}} \bm{\epsilon}_\theta \left(\hat{\bm{x}}_t, t, \bm{c}\right) \right) + \sigma_t\bm{z}$
					\ENDFOR
				\end{algorithmic}
				\textbf{return} $\hat{\bm{x}}_{0}$
			\end{algorithm}

		\subsubsection{Conditional DDIM}
			Since DDPM relies on a Markov chain for synthesis, it typically requires many steps ($T$ = 1000) to generate high-quality samples.
			Meanwhile, the number of synthesis steps must be the same as that in the training of DDPM.
			These drawbacks reduce the efficiency and flexibility of DDPM.

			To address the drawbacks of DDPM, DDIM \cite{generative_ddim} focuses on faster synthesis with fewer steps.
			DDIM generalizes DDPM with non-Markov diffusion processes that lead to the same training objective.
			Therefore, DDIM uses the same training algorithm as DDPM, and the trained $\bm{\epsilon}_\theta\left(\cdot\right)$ in DDPM can be reused by DDIM.
			Specifically, DDIM can use $T_S$ synthesis steps which can be fewer than $T$ in training $\bm{\epsilon}_\theta\left(\cdot\right)$.
			Given a trained $\bm{\epsilon}_\theta\left(\cdot\right)$, conditional DDIM synthesizes $\hat{\bm{x}}_0$ from $\hat{\bm{x}}_T\sim \mathcal{N}\left(\textbf{0}, \mathbf{I}\right)$  with $t\in[T_S,...,1]$ as:
			\begin{equation}
				\label{equ_conditional_ddim_synthesis}
				\begin{aligned}
					\hat{\bm{x}}_{t-1} = &\sqrt{\bar{\alpha}_{t-1}} \left(\frac{\hat{\bm{x}}_{t}-\sqrt{1-\bar{\alpha}_t}\cdot\bm{\epsilon}_\theta\left(\hat{\bm{x}}_t,t,\bm{c}\right)}{\sqrt{\bar{\alpha}_t}}\right)\\
					&+ \sqrt{1-\bar{\alpha}_{t-1}-\sigma_t^2}\cdot\bm{\epsilon}_\theta\left(\hat{\bm{x}}_t,t,\bm{c}\right) + \sigma_t \bm{z},
				\end{aligned}
			\end{equation}
			where $\sigma_t = \eta\sqrt{\left(1-\bar{\alpha}_{t-1}\right)/\left(1-\bar{\alpha}_{t}\right)}\sqrt{\beta_t}$ is the variance controlled by the factor $\eta$.

			In GenHPE, we use conditional DDIM with $T_S$ = 50 or $T_S$ = 100 under the conditions of $\bm{c}=\varphi\left(\bm{h}\right)$.
			Meanwhile, we use $\eta=0$ for deterministic synthesis.
			Algorithm \ref{algorithm_3} illustrates the synthesis with conditional DDIM in GenHPE.

			\begin{algorithm}[t]
				\caption{Synthesis with Conditional DDIM \cite{generative_ddim}}
				\label{algorithm_3}
				\textbf{Input:} ground-truth or manipulated skeleton vectors $\bm{h}$ 
				
				\begin{algorithmic}[1] 
					\STATE sample $\hat{\bm{x}}_T \sim \mathcal{N}\left(\textbf{0}, \textbf{I}\right)$
					\STATE calculate skeleton embeddings $\bm{c} = \varphi\left(\bm{h}\right)$
					\FOR {$t = T_S, ..., 1$}
					\STATE $\bm{z} \sim \mathcal{N}\left(\textbf{0}, \textbf{I}\right)$
					\STATE $\sigma_t = \eta\sqrt{\left(1-\bar{\alpha}_{t-1}\right)/\left(1-\bar{\alpha}_{t}\right)}\sqrt{\beta_t}$
					\STATE $\textstyle\begin{aligned}
						\textstyle\hat{\bm{x}}_{t-1} = &\sqrt{\bar{\alpha}_{t-1}} {\left(\hat{\bm{x}}_{t}-\sqrt{1-\bar{\alpha}_t}\cdot\bm{\epsilon}_\theta\left(\hat{\bm{x}}_t,t,\bm{c}\right)\right)}/{\sqrt{\bar{\alpha}_t}}\\
						&+ \sqrt{1-\bar{\alpha}_{t-1}-\sigma_t^2}\cdot\bm{\epsilon}_\theta\left(\hat{\bm{x}}_t,t,\bm{c}\right) + \sigma_t \bm{z}
					\end{aligned}$
					\ENDFOR
				\end{algorithmic}
				\textbf{return} $\hat{\bm{x}}_{0}$
			\end{algorithm}

		\subsubsection{CGAN}
			GAN \cite{generative_gan} designs a minimax two-player game to optimize a generator $g_\theta(\cdot)$ and a discriminator $f_\omega(\cdot)$ by adversarial learning.
			$g_\theta(\cdot)$ takes noise $\bm{z}\sim\mathcal{N}\left(\textbf{0}, \textbf{I}\right)$ as input to synthesize samples $g_\theta(\bm{z})$, which implicitly defines a generative distribution $p_g(\bm{z})$.
			$f_\omega(\cdot)$ takes samples $\bm{x}$ as input and outputs the probability $f_\omega(\bm{x})$ that $\bm{x}$ comes from ground-truth distribution $p_{\mathrm{gt}}(\bm{x})$ rather than $p_g(\bm{x})$.
			$f_\omega(\cdot)$ aims to distinguish $\bm{x}\sim p_{\mathrm{gt}}(\bm{x})$ from $\bm{z}\sim p_g(\bm{z})$, while $g_\theta(\cdot)$ aims to maximize the similarity between $ p_g(\bm{z})$ and $p_{\mathrm{gt}}(\bm{x})$.
			The distance between $p_g(\bm{z})$ and $p_{\mathrm{gt}}(\bm{x})$ can be represented by the Jensen–Shannon (JS) divergence $D_{\mathrm{JS}}$ as:
			\begin{equation}
				\label{equ_gan_js}
				\begin{aligned}
					\hspace{-2mm}D_{\mathrm{JS}}\left(p_{\mathrm{gt}} \| p_g\right) = \frac{1}{2}D_{\mathrm{KL}}\left(p_{\mathrm{gt}} \| p_{\mathrm{m}}\right)
					+ \frac{1}{2}D_{\mathrm{KL}}\left(p_g \| p_{\mathrm{m}}\right),
				\end{aligned}
			\end{equation}
			where $p_{\mathrm{m}}=\frac{p_{\mathrm{gt}}+p_g}{2}$.
			$D_{\mathrm{JS}}$ can be further formulated as an objective function:
			\begin{equation}
				\label{equ_gan_loss}
				\begin{aligned}
					\mathcal{L}\left(\theta, \omega\right) := &\mathbb{E}_{\bm{x}} \left[\log f_\omega\left(\bm{x}\right)\right] \\
					+ & \mathbb{E}_{\bm{z}} \left[\log \left(1-f_\omega\left(g_\theta\left(\bm{z}\right)\right)\right)\right].
				\end{aligned}
			\end{equation}
			The discriminator and the generator are optimized alternately with $\min_\theta\max_\omega\mathcal{L}\left(\theta, \omega\right)$, where $f_\omega(\cdot)$ aims to maximize the difference between its outputs $f_\omega\left(\bm{x}\right)$ from ground-truth inputs and its outputs $f_\omega\left(g_\theta\left(\bm{z}\right)\right)$ from synthetic inputs.
			On the contrary, $g_\theta\left(\cdot\right)$ aims to minimize their difference.
			However, this objective function lacks of continuity and usually leads to unstable training and mode collapse.
			Therefore, Wasserstein loss \cite{generative_wgan} has been proposed to solve these issues, which can be formulated as:
			\begin{equation}
				\label{equ_wgan_loss}
				\begin{aligned}
					\mathcal{L}\left(\theta, \omega\right) := \mathbb{E}_{\bm{x}} \left[f_\omega\left(\bm{x}\right)\right] - \mathbb{E}_{\bm{z}} \left[f_\omega\left(g_\theta\left(\bm{z}\right)\right)\right].
				\end{aligned}
			\end{equation}

			Original GAN and WGAN are unconditional, while a common practice of conditional GAN (CGAN) is to feed conditions into $g_\theta\left(\cdot\right)$ and $f_\omega\left(\cdot\right)$.
			Accordingly, the objective function of CGAN can be formulated as:
			\begin{equation}
				\label{equ_wcgan_loss}
				\begin{aligned}
					\mathcal{L}\left(\theta, \omega\right) := \mathbb{E}_{\bm{x}}\left[f_\omega\left(\bm{x}, \bm{c}\right)\right] - \mathbb{E}_{\bm{z}}\left[f_\omega\left(g_\theta\left(\bm{z}, \bm{c}\right), \bm{c}\right)\right],
				\end{aligned}
			\end{equation}
			In practice, CGAN is optimized by gradient descent, and thus $\max_\omega\mathcal{L}\left(\theta, \omega\right)$ is equivalent to perform gradient descent on $\triangledown_\omega \mathbb{E}_{\bm{x}, \bm{z}} \left[-f_\omega\left(\bm{x}, \bm{c}\right) + f_\omega\left(g_\theta\left(\bm{z}, \bm{c}\right), \bm{c}\right)\right]$.
			$\min_\theta\mathcal{L}\left(\theta, \omega\right)$ is equivalent to perform gradient descent on $\triangledown_\theta\mathbb{E}_{\bm{z}}\left[-f_\omega\left(g_\theta\left(\bm{z}, \bm{c}\right), \bm{c}\right)\right]$.

			\begin{algorithm}[t]
				\caption{Training Conditional GAN \cite{generative_cgan}}
				\label{algorithm_4}
				\textbf{repeat}
				\begin{algorithmic}[1] 
					\STATE sample ground-truth 3D HPE labels $\bm{y}$
					\STATE obtain ground-truth RF signal samples $\bm{x} \sim q\left(\bm{x}|\bm{y}\right)$
					\STATE convert $\bm{y}$ to skeleton vectors $\bm{h}$
					\STATE calculate skeleton embeddings $\bm{c} = \varphi\left(\bm{h}\right)$
					\FOR {discriminator steps}
					\STATE sample $\bm{z} \sim \mathcal{N}\left(\textbf{0}, \textbf{I}\right)$
					\STATE take gradient descent steps on \\
					\quad $\triangledown_\omega \mathbb{E}_{\bm{x}, \bm{z}} \left[-f_\omega\left(\bm{x}, \bm{c}\right) + f_\omega\left(g_\theta\left(\bm{z}, \bm{c}\right), \bm{c}\right)\right]$
					\ENDFOR
					\STATE take gradient descent steps on \\
					\qquad $\triangledown_\theta\mathbb{E}_{\bm{z}}\left[-f_\omega\left(g_\theta\left(\bm{z}, \bm{c}\right), \bm{c}\right)\right]$ 
				\end{algorithmic}
				\textbf{until} converged
			\end{algorithm}

			In GenHPE, we use skeleton embeddings $\bm{c}=\varphi\left(\bm{h}\right)$ as conditions.
			After the training of CGAN, we can directly employ $\hat{\bm{x}}=g_\theta\left(\bm{z}, \bm{c}\right)$ to synthesize RF signals from noise $\bm{z}$.
			Algorithm \ref{algorithm_4} presents the training of CGAN in GenHPE.

		\subsubsection{Backbone of Generative Models}
			We use a simplified ResNet \cite{exp_resnet} as the backbone of all generative models (\textit{i.e.}, $\bm{\epsilon}_\theta(\cdot)$ in diffusion models and $g_\theta(\cdot)$ in CGANs).
			This backbone has three blocks, and each block includes two convolutional layers.
			The numbers of filters in these layers over three blocks are \{256, 512, 256\}.
			The kernel sizes of convolutions over three blocks are \{11, 7, 5\}.
			The backbone of discriminator $f_\omega(\cdot)$ is implemented by an one-layer CNN whose convolutional layer has 64 filters with the kernel size of 3.
			All convolutional layers are followed by LeakyReLU activation functions.

	\subsection{Encoder-decoder Models}
		\label{appendix_encoder_decoder}
		GenHPE implements an encoder-decoder model to estimate the coordinates of human body joints in 3D space from RF signals.
		Herein, we describe the architecture of encoder-decoder models, as shown in Supplement Figure \ref{figure_encoder} and \ref{figure_decoder}.

		\subsubsection{Encoder}
\begin{figure}[t]
				\centering
				\includegraphics[width=0.80\linewidth]{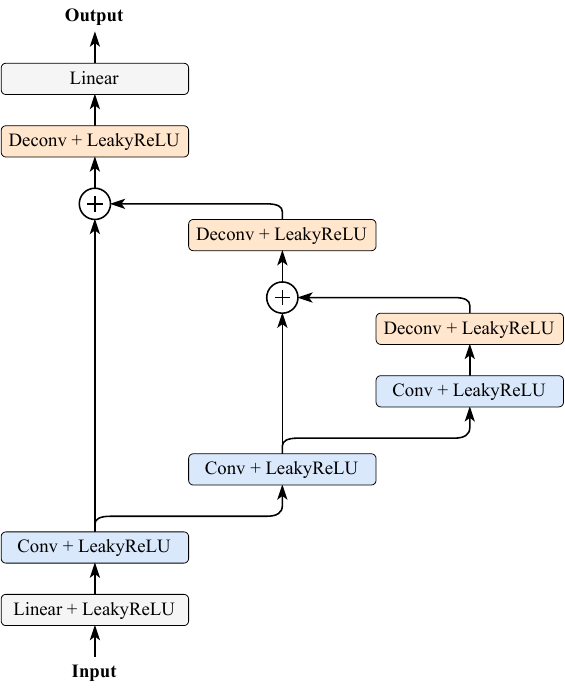}
				\caption{%
					The network architecture of encoder.
				}
				\label{figure_encoder}
			\end{figure}

			We design a simplified U-Net model \cite{exp_unet} as the backbone of encoder.
			Supplement Figure \ref{figure_encoder} presents the network architecture of encoder, including three convolutional (Conv) layers and three deconvolutional (Deconv) layers.
			Specifically, each convolutional layer has 256 filters, while the kernel sizes are \{7, 5, 3\} for three layers, respectively.
			On the contrary, the kernel sizes of three deconvolutional layers are \{3, 5, 7\}, while each layer also has 256 filters.
			%
		
		%
		\subsubsection{Decoder}
\begin{figure}[t]
			\centering
			\includegraphics[width=0.80\linewidth]{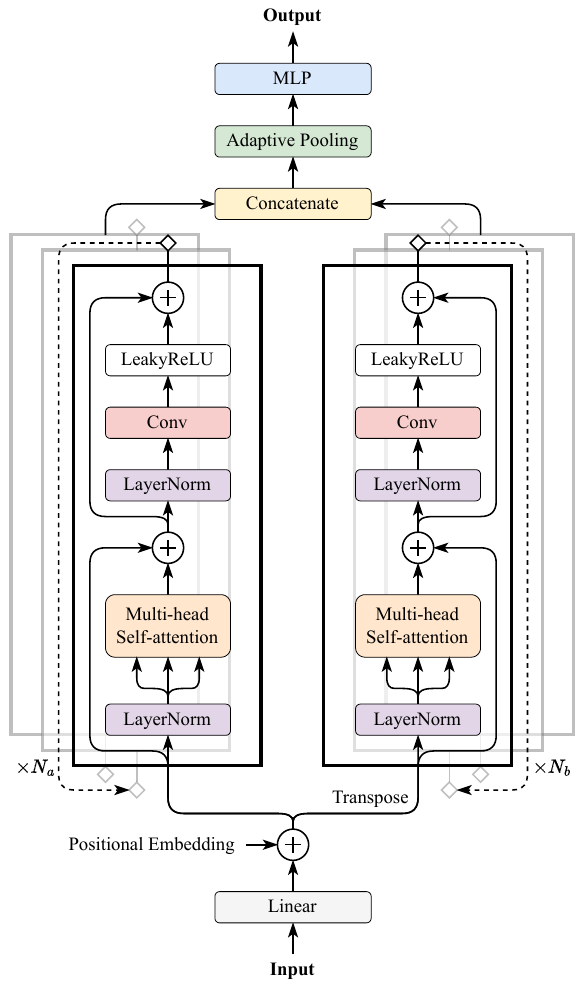}
			\caption{%
				The network architecture of decoder.
			}
			\label{figure_decoder}
		\end{figure}

			We devise a two-stream decoder with self-attention layers augmented by convolutional layers.
			Supplement Figure \ref{figure_decoder} presents the network architecture of decoder.
			Two streams have $N_a$ and $N_b$ blocks, respectively.
			Specifically, the first stream has 8 blocks ($N_a$ = 8), where each self-attention layer has 8 heads, and the kernel sizes of convolutional layers over 8 blocks are \{9, 7, 5, 3, 9, 7, 5, 3\} with the dilation \{2, 1, 2, 1, 2, 1, 2, 1\}.
			The second stream takes transposed signals as input to model feature dependencies from another view, including 4 blocks ($N_b$ = 4) with 4 heads in each self-attention layer.
			In the second stream, the kernel sizes of convolutional layers over 4 blocks are \{3, 3, 3, 3\} with the dilation \{1, 1, 1, 1\}.
			We concatenate the features from two streams and apply adaptive pooling to reduce feature dimensions.
			Finally, the features are fed into a multilayer perceptron (MLP) with two linear layers to output pose estimations, and the first linear layer is followed by a LeakyReLU activation function.

\section{Experiments}
	Herein, we provide more details about our experiments, as well as more quantitative analysis and visualizations.

	\subsection{Datasets}
		\label{appendix_exp_dataset}
		Our experiments include three public datasets, which collect different sources of RF signals, including WiFi, ultra-wideband (UWB), and millimeter wave (mmWave).
		Each dataset provides a set of RF signal samples and the corresponding annotations of human body joints in 3D space.
		
		%
		\subsubsection{WiFi}
		The WiFi dataset \cite{method_piw_3d} collects WiFi Channel State Information (CSI) using the carrier frequency of 5.64 GHz.
		The dimensions of each CSI sample ($\bm{x}$ in our paper) are 60$\times$180, annotated by a human pose label ($\bm{y}$ in our paper) whose dimensions are 14$\times$3 (\textit{i.e.}, 14 human body joints in 3D space).
		More descriptions about the dimensions of samples and labels are provided in the paper of this dataset \cite{method_piw_3d}.

		There are 7 subjects and 3 environments in this dataset, from which we extract 30703 CSI samples for evaluation.
		(1) For Random splits, there are 24565 samples in the training set, 3071 samples in the validation set, and 3071 samples in the test set.
		(2) For Cross-Subject splits, there are 24746, 2750, and 3211 samples in the training, validation, and test sets, respectively.
		(3) For Cross-Environment splits, the training set includes 18914 samples, the validation set has 2102 samples, and the test set contains 9691 samples.
		
		%
		\subsubsection{UWB}
		The UWB dataset \cite{data_operanet} uses 3.99$\sim$4.49 GHz signals to collect RF signal samples, involving 6 subjects and 2 environments.
		Each UWB sample contains 35 channels and 40 time steps under the sample rate of 195$\sim$400 Hz, and each sample is annotated by 19 human body joints in 3D space.
		The dimensions of each UWB sample ($\bm{x}$ in our paper) are 70$\times$40, including both the amplitudes and phases, while the dimensions of each human pose label ($\bm{y}$ in our paper) are 19$\times$3.
		More descriptions about the dimensions of samples and labels are provided in the paper of this dataset \cite{data_operanet}.

		We extract 293930 samples from this dataset for evaluation.
		(1) For Random splits, we use 118150 samples as the training set, 14769 samples as the validation set, and 14769 samples as the test set.
		(2) For Cross-Subject splits, the numbers of samples are 101205, 11246, and 33791 for the training set, validation set, and test set, respectively.
		(3) For Cross-Environment splits, 53504, 5945, and 86793 samples are adopted for training, validation, and testing, respectively. 
		
		%
		\subsubsection{mmWave}
		The mmWave dataset \cite{data_mmfi} collects RF signals with 6 subjects and 4 environments using the signal frequency of 60$\sim$64 GHz.
		We extract 95666 samples by combining multiple frames in each 0.5-second window, and each sample includes a point cloud with the label of 17 human body joints in 3D space.
		In the point cloud sample, each point has 5 dimensions indicating its 3D coordinates, Doppler velocity, and signal intensity.
		Since the maximum number of points is 493, the dimensions of each point cloud sample ($\bm{x}$ in our paper) are 5$\times$493, and the dimensions of each human pose label ($\bm{y}$ in our paper) are 17$\times$3.
		More descriptions about the dimensions of samples and labels are provided in the paper of this dataset \cite{data_mmfi}.

		(1) For Random splits, the training set includes 76532 samples, the validation set has 9567 samples, and the test set contains 9567 samples.
		(2) For Cross-Subject splits, there are 66830, 7426, and 21410 samples in the training, validation, and test sets, respectively.
		(3) For Cross-Environment splits, we employ 63373, 7042, and 25251 samples for training, validation, and testing, respectively.

	\subsection{Evaluation Metrics}
		\label{appendix_exp_metrics}
		\noindent\textbf{Mean Per Joint Position Error (MPJPE).}
		MPJPE computes the Euclidean distance between ground-truth and estimated human poses.
		Given a ground-truth label of $N$ human body joints $\bm{y}\in\mathbb{R}^{N\times 3}$, the MPJPE of model estimation $\hat{\bm{y}}$ can be calculated as:
		\begin{equation}
			\label{equ_mpjpe}
			\begin{aligned}
				\mathrm{MPJPE}\left(\hat{\bm{y}}, \bm{y}\right) = &\frac{1}{N}\sum_{i=1}^{N}\left\| \hat{\bm{y}}_i-\bm{y}_i \right\|_2\\
				= &\frac{1}{N}\sum_{i=1}^{N}\sqrt{\sum_{j=1}^{3}\left(\hat{\bm{y}}_{i,j} - \bm{y}_{i,j}\right)^2},
			\end{aligned}
		\end{equation}
		where $i$ is the joint index and $j$ is the dimension index.

		\begin{figure*}[t]
			\centering
			\begin{subfigure}{\linewidth}
				\centering
				\includegraphics[width=\linewidth]{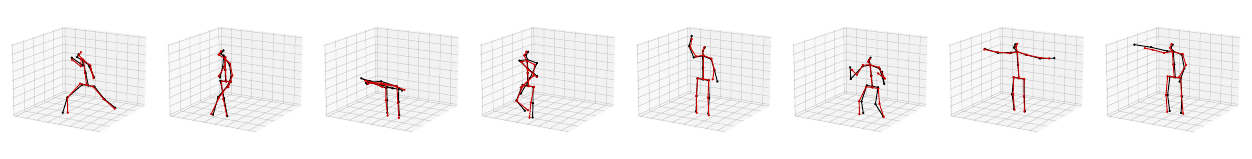}
				\caption{Performance of GenHPE with generative counterfactuals using DDPMs ($T$ = 1000).}
				\label{figure_mmfi_cross_subject_ddpm_supp}
			\end{subfigure}
			
			\begin{subfigure}{\linewidth}
				\centering
				\includegraphics[width=\linewidth]{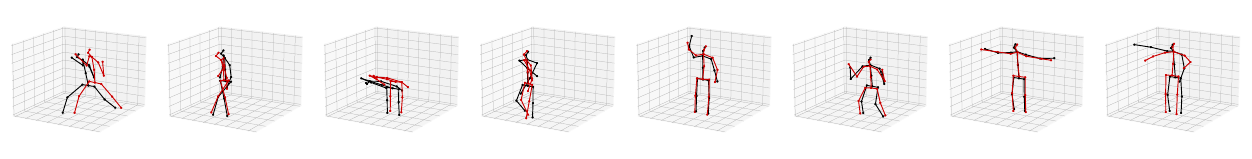}
				\caption{Performance of GenHPE without generative counterfactuals.}
				\label{figure_mmfi_cross_subject_nogen_supp}
			\end{subfigure}
			\caption{%
				Performance of GenHPE with/without generative counterfactuals in cross-subject scenarios.
				Black: ground-truth human poses. \textcolor[HTML]{CC0000}{Red}: model estimations.
			}
			\label{figure_mmfi_cross_subject_supp}
		\end{figure*}%

		\begin{figure*}[t]
			\centering
			\begin{subfigure}{\linewidth}
				\centering
				\includegraphics[width=\linewidth]{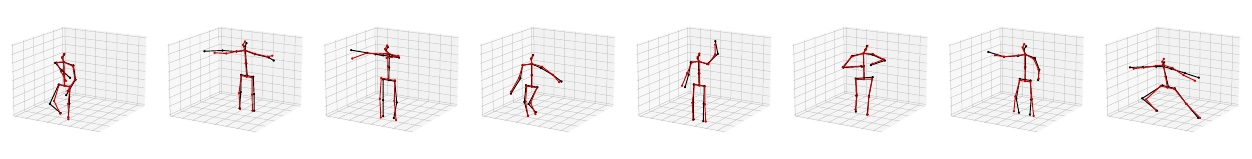}
				\caption{Performance of GenHPE with generative counterfactuals using DDPMs ($T$ = 1000).}
				\label{figure_mmfi_cross_env_ddpm_supp}
			\end{subfigure}
			
			\begin{subfigure}{\linewidth}
				\centering
				\includegraphics[width=\linewidth]{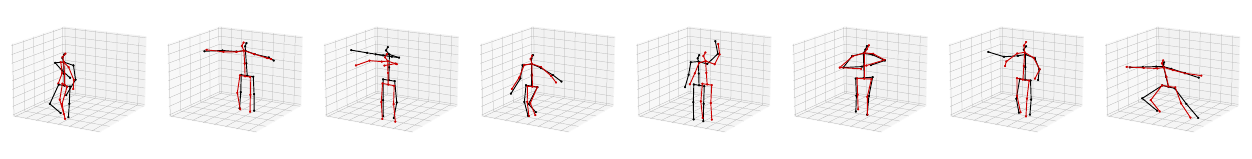}
				\caption{Performance of GenHPE without generative counterfactuals.}
				\label{figure_mmfi_cross_env_nogen_supp}
			\end{subfigure}
			\caption{%
				Performance of GenHPE with/without generative counterfactuals in cross-environment scenarios.
				Black: ground-truth human poses. \textcolor[HTML]{CC0000}{Red}: model estimations.
			}
			\label{figure_mmfi_cross_env_supp}
		\end{figure*}

		\noindent\textbf{Procrustes-aligned MPJPE (PA-MPJPE).}
		PA-MPJPE \cite{exp_pa_mpjpe} adopts Procrustes alignment to transform the estimated pose $\hat{\bm{y}}$ into $\hat{\bm{y}}'$ according to the similarity between $\hat{\bm{y}}$ ann $\bm{y}$.
		PA-MPJPE calculates the MPJPE of transformed pose estimation $\hat{\bm{y}}'$ as:
		\begin{equation}
			\label{equ_pa_mpjpe}
			\begin{aligned}
				\mathrm{PA\textendash MPJPE}\left(\hat{\bm{y}}, \bm{y}\right) = \mathrm{MPJPE}\left(\hat{\bm{y}}', \bm{y}\right).
			\end{aligned}
		\end{equation}

		\noindent\textbf{Mean Per Joint Dimension Location Error (MPJDLE).}
		MPJDLE \cite{method_piw_3d} measures the mean absolute error of estimated pose $\hat{\bm{y}}$ in each dimension as:
		\begin{equation}
			\label{equ_mpjdle}
			\begin{aligned}
				\mathrm{MPJDLE}_j\left(\hat{\bm{y}}, \bm{y}\right) = \frac{1}{N}\sum_{i=1}^{N}\left|\hat{\bm{y}}_{i,j} - \bm{y}_{i,j}\right|,
			\end{aligned}
		\end{equation}
		and the overall MPJDLE of $\hat{\bm{y}}$ can be calculated as $\mathrm{MPJDLE}\left(\hat{\bm{y}}, \bm{y}\right) = \frac{1}{3}\sum_{j=1}^{3}\mathrm{MPJDLE}_j$.

	\subsection{Hyper-parameter Sensitivity Study}
		\label{appendix_exp_sensitivity}
		We further evaluate the impact of factor $\lambda$ on the performance of GenHPE. 
		Specifically, we vary $\lambda\in[0.0, 0.2, 0.4, 0.6, 0.8, 1.0]$ with WiFi and UWB in cross-subject and cross-environment scenarios.

		As shown in Supplement Table \ref{table_sensitive_piw_3d}, GenHPE consistently has lower estimation errors when $\lambda>0.0$, which demonstrates the efficacy of counterfactual regularization for domain-independent representation learning with WiFi.
		When we set $0.2\leq\lambda\leq1.0$, GenHPE shows stable performance for 3D HPE, proving its robustness for domain generalization with varying values of $\lambda$.

		Supplement Table \ref{table_sensitive_operanet} discusses the performance of GenHPE with UWB using different values of $\lambda$.
		Results demonstrate that GenHPE achieves robust and state-of-the-art domain generalization performance for 3D HPE when $\lambda>0.0$.

	\subsection{Supplemental Visualizations}
		We discuss the effectiveness of generative counterfactuals in GenHPE with mmWave.
		Supplement Figure \ref{figure_mmfi_cross_subject_supp} and \ref{figure_mmfi_cross_env_supp} visualize 3D HPE results with/without generative counterfactuals in cross-subject and cross-environment scenarios.
		Results intuitively illustrate that GenHPE achieves more precise 3D HPE by learning domain-independent representations for cross-domain 3D HPE.

		\begin{table}[t]
			\footnotesize 
			\centering
			\begin{tabular}{cc@{\quad}c@{\quad}c@{\quad}c}
				
				\toprule
				
				&$\lambda$	& MPJPE	& PA-MPJPE 		& MPJDLE 	 	\\ 
				
				\midrule
				& 0.0 	& 260.59\std{4.43}		& 112.64\std{1.13}			& 122.62\std{2.15}		\\	
				& 0.2 	& 209.53\std{1.92}		& \ \ 94.86\std{2.23}	& 100.80\std{1.15}		\\	
				Cross-			& 0.4 	& 211.64\std{2.84}		& \ \ 95.22\std{1.77}	& 101.87\std{1.68}		\\	
				Subject			& 0.6 	& 211.64\std{2.82}		& \ \ 95.21\std{1.75}	& 101.87\std{1.67}		\\	
				& 0.8 	& 211.64\std{2.82}		& \ \ 95.21\std{1.74}	& 101.86\std{1.66}		\\	
				& 1.0 	& 211.81\std{2.84}		& \ \ 95.21\std{1.74}	& 101.94\std{1.67}		\\	
				\midrule
				& 0.0 	& \ \ 261.86\std{12.97}	& 117.22\std{7.80}	& 126.25\std{8.01}		\\
				& 0.2 	& 230.30\std{2.46}		& 106.71\std{0.81}	& 109.26\std{0.92}		\\	
				Cross-			& 0.4 	& 229.62\std{2.93}		& 106.09\std{0.99}	& 108.87\std{1.43}		\\	
				Environment		& 0.6 	& 228.27\std{1.33}		& 106.04\std{0.82}	& 108.08\std{0.80}		\\	
				& 0.8 	& 228.18\std{0.75}		& 105.79\std{0.58}	& 108.01\std{0.30}		\\	
				& 1.0 	& 228.88\std{0.56}		& 105.81\std{0.90}	& 108.17\std{0.24} 		\\	
				\bottomrule
				
			\end{tabular}
			\caption{Performance of GenHPE regarding different regularization factors $\lambda$ for 3D HPE with WiFi \cite{method_piw_3d}. (Unit: mm)}
			\label{table_sensitive_piw_3d}
		\end{table}

		\begin{table}[t]
			\footnotesize 
			\centering
			\begin{tabular}{cc@{\quad}c@{\quad}c@{\quad}c}
				
				\toprule
				
				&$\lambda$	& MPJPE 	& PA-MPJPE 		& MPJDLE 	 	\\ 
				
				\midrule
				& 0.0 	& 298.99\std{2.61}	& 135.73\std{0.45}	& 138.81\std{1.00}		\\
				& 0.2 	& 279.14\std{5.13}	& 125.22\std{1.42}	& 128.89\std{1.82}		\\	
				Cross-			& 0.4 	& 277.37\std{3.28}	& 125.23\std{1.32}	& 128.27\std{1.18}		\\	
				Subject			& 0.6 	& 279.91\std{5.97}	& 125.30\std{1.06}	& 129.15\std{2.21}		\\	
				& 0.8 	& 278.29\std{2.61}	& 125.12\std{0.62}	& 128.44\std{1.02}		\\
				& 1.0 	& 279.17\std{3.99}	& 124.42\std{1.10}	& 128.79\std{1.46}		\\	
				\midrule
				& 0.0 	& 337.45\std{3.45}		& 146.74\std{0.49}	& 153.21\std{1.14}		\\
				& 0.2 	& \ \ 295.37\std{10.39}	& 135.48\std{6.27}	& 136.49\std{4.76}		\\	
				Cross-			& 0.4 	& 292.70\std{9.57}		& 134.00\std{5.54}	& 135.18\std{4.36}		\\	
				Environment		& 0.6 	& 290.32\std{5.91}		& 132.38\std{2.98}	& 133.93\std{2.45}		\\	
				& 0.8 	& \ \ 293.19\std{10.16}	& 133.73\std{4.62}	& 135.41\std{4.50}		\\	
				& 1.0 	& \ \ 292.80\std{12.66}	& 133.97\std{6.52}	& 135.36\std{5.84}		\\
				\bottomrule
			\end{tabular}
			\caption{Performance of GenHPE regarding different regularization factors $\lambda$ for 3D HPE with UWB \cite{data_operanet}. (Unit: mm)}
			\label{table_sensitive_operanet}
		\end{table}

	{
	    \small
	    \bibliographystyle{ieeenat_fullname}
	    \bibliography{main}
	}

\end{document}